\documentclass[10pt, letter, onecolumn]{arxiv}

\usepackage{kantlipsum, lipsum}
\usepackage{dm-colors}
\usepackage{amsmath}
\usepackage{pstricks, pst-node}
\usepackage{verbatim}
\usepackage{multirow}
\usepackage{array}
\usepackage{longtable}
\usepackage{pdflscape}
\usepackage{scalerel}
\usepackage{booktabs}
\usepackage{enumitem}
\usepackage{xspace}
\usepackage{bm}
\usepackage{bbm}
\usepackage{mathtools}
\usepackage{soul}
\usepackage{epsfig}
\usepackage{graphicx}
\usepackage{amssymb}
\usepackage{colortbl}
\usepackage{csquotes}
\usepackage{setspace}
\usepackage{colortbl}
\usepackage{bbding}
\usepackage{threeparttable}
\usepackage{tabularx,ragged2e}
\usepackage{placeins}
\usepackage{bbding}
\definecolor{darkpastelgreen}{rgb}{0.13, 0.55, 0.13}
\definecolor{darkpastelred}{rgb}{0.55, 0.13, 0.13}
\definecolor{mygray}{rgb}{1, 1, 1}
\usepackage{lmodern}
\usepackage[hang,flushmargin]{footmisc}

\usepackage{nameref}
\usepackage{varioref}
\usepackage{amssymb}
\usepackage{pifont}
\usepackage{rotating}
\usepackage{graphicx}

\usepackage[pagebackref=false,breaklinks=false,%
            colorlinks=true,bookmarks=true,citecolor=ourdarkblue,%
            urlcolor=ourdarkblue,linkcolor=ourdarkblue]{hyperref}
\usepackage[noabbrev,capitalize]{cleveref}
\usepackage{etoc}
\usepackage{lineno}
\usepackage{algorithm}
\usepackage{makecell} 
\usepackage{algpseudocode}

\usepackage{amsmath}
\usepackage{thmtools}
\usepackage{tcolorbox}
\usepackage{lmodern}

\declaretheoremstyle[
    spaceabove=6pt, spacebelow=6pt,
    headfont=\bfseries, headpunct={.}, headformat={\NAME\ \NUMBER},
    bodyfont=\normalfont,
    postheadspace=0.5em
]{promptstyle}

\declaretheorem[name=Prompt, style=promptstyle]{prompt}

\tcolorboxenvironment{prompt}{
    colback=gray!10!,
    colframe=gray!75!,
    fonttitle=\bfseries,
    title=Prompt,
    boxrule=0.5pt,
    sharp corners
}

\graphicspath{{figures/}}
\definecolor{mygray}{rgb}{0.85, 0.85, 0.85}

\title{\Large{Quantifying the Reasoning Abilities of LLMs \\ on  Real-world Clinical Cases}}

\author[1,2,$\ast$]{Pengcheng Qiu} 
\author[1,2,$\ast$]{Chaoyi Wu} 
\author[1]{Shuyu Liu}
\author[1,2]{Weike Zhao}
\author[3]{Zhuoxia Chen}
\author[3]{\\ \vspace{0.1cm}Hongfei Gu}
\author[3]{Chuanjin Peng}
\author[1,2]{Ya Zhang}
\author[1,2,$\dag$]{Yanfeng Wang} 
\author[1,2,$\dag$]{Weidi Xie}

\affil[1]{\normalsize Shanghai Jiao Tong University, Shanghai, China \authorcr \vspace{0.1cm}}

\affil[2]{\normalsize Shanghai Artificial Intelligence Laboratory, Shanghai, China  \authorcr \vspace{0.1cm}}

\affil[3]{\normalsize China Mobile Communications Group Shanghai Co., Ltd., Shanghai, China \authorcr}

\affil[$\ast$]{\normalsize Equal contributions\hspace{1cm}}
\affil[$\dag$]{\normalsize Corresponding author\authorcr Yanfeng Wang: wangyanfeng622@sjtu.edu.cn; Weidi Xie: weidi@sjtu.edu.cn}

\begin{document}

\begin{abstract}
Recent advancements in reasoning-enhanced large language models (LLMs), such as DeepSeek-R1 and OpenAI-o3, have demonstrated significant progress. However, their application in professional medical contexts remains underexplored, particularly in evaluating the quality of their reasoning processes alongside final outputs.
Here, we introduce \textbf{MedR-Bench}, a benchmarking dataset of 1,453 structured patient cases, annotated with reasoning references derived from clinical case reports. 
Spanning 13 body systems and 10 specialties, it includes both common and rare diseases. 
To comprehensively evaluate LLM performance, we propose a framework encompassing three critical \textbf{examination recommendation}, \textbf{diagnostic decision-making}, 
and \textbf{treatment planning}, simulating the entire patient care journey.
To assess reasoning quality, we present the \textbf{Reasoning Evaluator}, 
a novel automated system that objectively scores free-text reasoning responses based on \textbf{efficiency}, \textbf{factuality}, and \textbf{completeness} using dynamic cross-referencing and evidence checks. Using this benchmark, we evaluate five state-of-the-art reasoning LLMs, including DeepSeek-R1, OpenAI-o3-mini, and Gemini-2.0-Flash Thinking, etc. 
Our results show that current LLMs achieve over 85\% accuracy in relatively simple diagnostic tasks when provided with sufficient examination results. 
However, performance declines in more complex tasks, such as examination recommendation and treatment planning. While reasoning outputs are generally reliable, with factuality scores exceeding 90\%, critical reasoning steps are frequently missed.
These findings underscore both the progress and limitations of clinical LLMs. Notably, open-source models like DeepSeek-R1 are narrowing the gap with proprietary systems, highlighting their potential to drive accessible and equitable advancements in healthcare.

\end{abstract}

\maketitle


\section{Introduction}

Large language models (LLMs) have advanced significantly in recent years, with systems such as OpenAI-o1~\cite{jaech2024openai} and DeepSeek-R1~\cite{guo2025deepseek} demonstrating remarkable reasoning capabilities. These models have excelled in structured problem-solving and logical inference, achieving notable success in fields like mathematics and programming~\cite{zhong2024evaluation, guo2025deepseek, phan2025humanity}. However, their application in the medical domain—a field defined by complexity, high stakes, and the need for contextual understanding—remains underexplored.

Existing medical LLM benchmarks~\cite{jin-etal-2019-pubmedqa, jin2021disease, pal2022medmcqa, wu2025towards, singhal2023large, singhal2025toward, wu2024pmc, qiu2024towards, xie2024preliminary, hager2024evaluation, lamparth2025moving} primarily focus on evaluating final outputs, such as diagnostic accuracy or treatment recommendations, without adequately assessing the reasoning processes behind them. This approach contrasts with clinical practice, where physicians construct logical reasoning chains by synthesizing diverse and often incomplete information to guide decision-making. Consequently, the lack of benchmarks prioritizing reasoning quality represents a significant gap in evaluating LLMs’ reliability and utility in real-world clinical settings.

To address this, we propose \textbf{MedR-Bench}, the first benchmark specifically designed to evaluate the medical reasoning capabilities of state-of-the-art LLMs. MedR-Bench includes 1,453 real-world clinical cases spanning 13 body systems and 10 disorder types, with 656 cases dedicated to rare diseases. Unlike existing benchmarks, MedR-Bench emphasizes not only the correctness of final diagnoses or treatment plans but also the \textbf{transparency}, \textbf{coherence}, and \textbf{factual soundness} of the reasoning processes behind them. 
Inspired by prior works~\cite{zhao2022pmc, wu2023towards}, 
the benchmark is constructed from real-world case reports in the PMC Open Access Subset~\cite{pmc_open_access_subset}, reorganized into structured patient cases using GPT-4o. 
Each case consists of (i) detailed patient information ({\em e.g.}, chief complaint, medical history), (ii) a structured reasoning process derived from case discussions, and (iii) the final diagnosis or treatment plan, reflecting real-world clinical complexity. By incorporating diverse and challenging cases, including rare conditions, MedR-Bench serves as a comprehensive testbed for assessing the reasoning capabilities of LLMs in clinical environments.

To evaluate LLMs, we propose a framework spanning three critical clinical stages: 
\textbf{examination recommendation}, \textbf{diagnostic decision-making}, and \textbf{treatment planning}, capturing the entire patient care trajectory. 
Examination recommendation evaluates the model’s ability to suggest relevant clinical assessments and iteratively gather necessary information. 
Diagnostic decision-making tests the model’s ability to derive precise diagnoses based on patient history, examination findings, lab tests and imaging findings. 
Finally, treatment planning assesses the model’s ability to recommend appropriate interventions, such as monitoring strategies, medications, or surgical options, grounded in diagnostic conclusions and patient context.

To quantify performance, we develop an evaluation system to assess both reasoning quality and final outputs. For reasoning evaluation, we introduce the \textbf{Reasoning Evaluator}, 
a novel automated system that validates free-text reasoning processes using web-scale medical resources and perform cross-referencing. It calculates metrics for \textbf{efficiency}, \textbf{factuality}, and \textbf{completeness}. For final outputs, we adopt standard metrics such as accuracy, precision, and recall. Using MedR-Bench, we evaluate five reasoning-enhanced LLMs—OpenAI-o3-mini, Gemini-2.0-Flash Thinking, DeepSeek-R1, Qwen-QwQ, and Baichuan-M1—providing a comparative analysis of their strengths and limitations across various clinical stages.

Our findings reveal that current clinical LLMs perform well on relatively simple tasks, 
such as generating accurate diagnoses when sufficient information is available, achieving over 85\% accuracy. However, they struggle with complex tasks, such as examination recommendation and treatment planning. In terms of reasoning quality, LLMs exhibit strong factual accuracy, with nearly 90\% of reasoning steps being correct, but omissions in critical reasoning steps are common,
indicating a need for improved reasoning completeness. For rare diseases, while these cases remain challenging, models generally show consistent performance across reasoning and prediction tasks, suggesting a robust understanding of medical knowledge across case types. 

Encouragingly, our findings suggest that open-source models, such as \textbf{DeepSeek-R1}, are steadily closing the gap with proprietary systems like \textbf{OpenAI-o3-mini},
underscoring their potential to drive accessible and equitable healthcare innovations, 
motivating continued efforts in their development. All codes, data, assessed model responses, and the evaluation pipeline are fully open-source in \href{https://github.com/MAGIC-AI4Med/MedRBench}{MedR-Bench}.

\begin{figure}
    \centering
    \includegraphics[width=\linewidth]{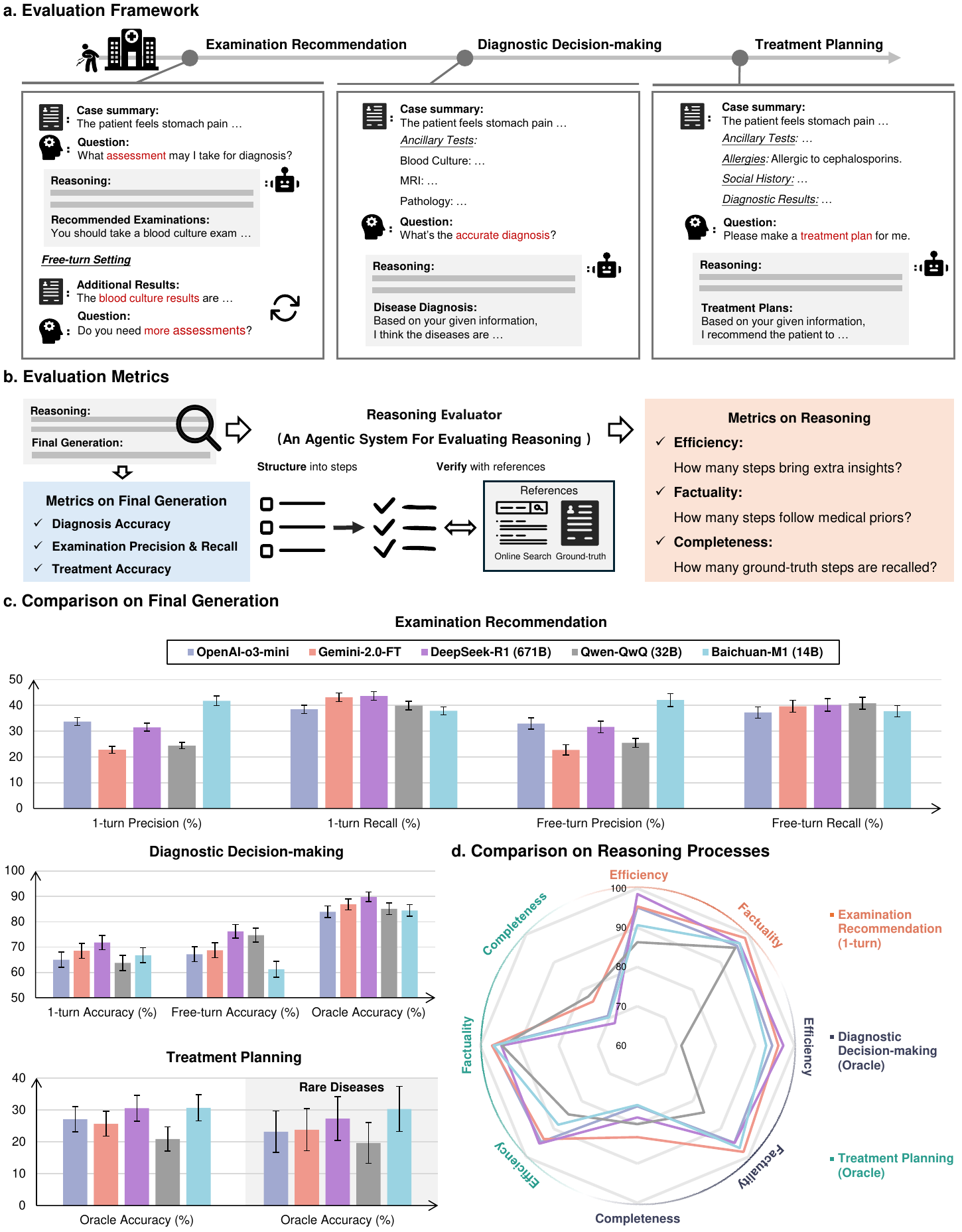}
    \caption{\textbf{Overview of our main evaluation pipeline and results}. \textbf{a} illustrates our evaluation framework across three critical patient stages. \textbf{b} presents the metrics for reasoning processes and final generations using our Reasoning Evaluator. \textbf{c} compares the performance of five LLMs on examination recommendation, diagnostic decision-making, and treatment planning. Notably, for treatment planning, we include a comparison on rare disease cases. For other settings, as the rare disease results show minimal variation compared to all cases, we omit them here and provide them in the extended tables. \textbf{d} compares the qualities of reasoning processes, with results for rare cases also provided in the supplementary material. For examination recommendation, 1-turn reasoning results are plotted, and for diagnostic decision, oracle reasoning results are plotted.}
    \label{fig:teaser}
\end{figure}

\section{Results}
In this section, we present our main findings. We begin with an overview of \textbf{MedR-Bench}, followed by an analysis of results across the three key stages: 
\textbf{examination recommendation}, \textbf{diagnostic decision-making}, and \textbf{treatment planning}. In Supplementary~\ref{case_study}, we provide qualitative case studies.

\subsection{LLMs Models For Evaluation}

This study utilizes a range of models with varying versions, sizes, cut-off dates for training data, and release dates. For closed-source models, we accessed their APIs directly, while for open-source models, we downloaded the model weights and conducted local inference. The details are presented below.

\vspace{-0.3cm}
\begin{itemize}
\setlength\itemsep{3pt}
    \item \textbf{OpenAI-o3-mini}: This is a closed-source model with the version identifier o3-mini-2025-01-31. Its model size is not disclosed. The cut-off date for training data is October 2023, and it was officially released in January 2025.
    
    \item \textbf{Gemini-2.0-FT}: 
    This is closed-source model, identified by the version gemini-2.0-flash-thinking-exp-01-21. Similar to OpenAI-o3-mini, the model size is not disclosed. Its cut-off date for training data is June 2024, and it was released in January 2025.

    \item \textbf{DeepSeek-R1}: 
    This is an open-source model with the version identifier deepseek-ai/DeepSeek-R1. 
    It is a large-scale model with 671 billion parameters (671B). 
    The cut-off date for training data is not disclosed, and it was released in January 2025.

    \item \textbf{Qwen-QwQ}: 
    This is an open-source model with the version identifier Qwen/QwQ-32B-Preview. 
    It has 32 billion parameters (32B). The cut-off date for training data is not disclosed, 
    and the model was released in November 2024.

    \item \textbf{Baichuan-M1}: 
    This is an open-source model with the version identifier baichuan-inc/Baichuan-M1-14B-Instruct. It has 14 billion parameters (14B), with no disclosed cut-off date for training data. 
    The model was released in January 2025.
    
\end{itemize}

\subsection{Introduction of MedR-Bench}
Our proposed \textbf{MedR-Bench} comprises three key components: 
(1) structured patient cases, (2) a versatile evaluation framework spanning three stages, and (3) a comprehensive set of evaluation metrics.

\subsubsection{Patient Cases}
Leveraging the case reports from the PMC Open Access Subset~\cite{pmc_open_access_subset}, 
we compiled a dataset of 1,453 patient cases published after July 2024 to ensure a fair and robust assessment across all models based on their cut-off date for training data.
These are divided into two subsets: 
\textbf{MedR-Bench-Diagnosis} with 957 diagnosis-related cases, 
and \textbf{MedR-Bench-Treatment} with 496 treatment-related cases. 
As illustrated in Supplementary Figure~\ref{fig:case1}, all cases are systematically organized into the following elements:

\vspace{-0.3cm}
\begin{itemize} 
\setlength\itemsep{3pt}
    \item \textbf{Case Summary}: Documents key patient information. 
    For diagnosis cases, this includes basic patient demographics ({\em e.g.}, age, sex), 
    chief complaint, history of present illness, past medical history, 
    family history, physical examination, and ancillary tests ({\em e.g.}, lab and imaging results). For treatment cases, additional factors such as allergies, social history, and diagnostic results are included, as these influence treatment decisions. Any missing information in the raw case reports is recorded as ``not mentioned''.

    \item \textbf{Reasoning Processes}: Summarized from the discussion sections of case reports, this captures the logical steps used to reach a diagnosis or formulate a treatment plan. For diagnosis cases, the reasoning focuses on methods like differential diagnosis. 
    For treatment cases, it emphasizes treatment goals and the rationale behind the chosen interventions.    
    
    \item \textbf{Diagnosis or Treatment Results}: Directly extracted from the raw case reports. For diagnosis, this includes identified diseases. For treatment, it consists of free-text descriptions of the recommended interventions.
\end{itemize}

\vspace{-0.3cm}
Additionally, each case is categorized by ``body system'' and ``disorders and conditions'' following the taxonomy from MedlinePlus\footnote{\url{https://medlineplus.gov/healthtopics.html}}, and cases are flagged as rare disease-related when applicable. 
This allows MedR-Bench-Diagnosis and MedR-Bench-Treatment to be further split to create rare disease subsets containing 491 and 165 cases, respectively. 
Case distributions are detailed in the \textbf{Methods} section, with patient case examples provided in Supplementary~\ref{case_study}.

\subsubsection{Evaluation Settings}

To evaluate LLMs’ clinical capabilities, we developed a framework covering three stages of the patient care journey: \textbf{examination recommendation}, \textbf{diagnostic decision-making}, and \textbf{treatment planning}, as shown in Figure~\ref{fig:teaser}a~(More detailed demonstrations are shown in Extended Figure~\ref{fig:settings_show}). 
Below, we summarize these components~(details are provided in the Methods section).

\textbf{Examination Recommendation}.

This setting simulates a scenario where a patient first visits a hospital, and LLMs are tasked with recommending examinations items such as lab tests or imaging studies, iteratively gathering information to aid diagnosis or treatment. 
Using the \textbf{MedR-Bench-Diagnosis}, the case summaries—excluding ancillary test results—serve as input, while the real-world ancillary test events are used as for ground-truth reference.
Similar to previous works~\cite{hager2024evaluation, johri2025evaluation, liao2024automatic}, 
we initialize an LLM-powered agent to play the role of the patient. The assessed clinical LLM can interact with this agent by recommending relevant examination items, and the agent provides corresponding results.

To evaluate performance, we define two sub-settings:
(i) 1-turn examination recommendation: LLMs can query examination results in a single round of interaction; (ii) Free-turn examination recommendation: LLMs can query information through multiple rounds until sufficient information is gathered for subsequent decisions.

\textbf{Diagnostic Decision-Making}.

This setting evaluates whether LLMs can deliver accurate diagnoses based on the given patient information. Using the \textbf{MedR-Bench-Diagnosis}, case summaries serve as input, 
while the recorded diagnoses serve as the ground truth. 

We define three sub-settings based on the availability of examination information:
(i) diagnostic decision after 1-turn examination recommendation: LLMs use the limited information gathered from the 1-turn setting;
(ii) diagnostic decision after free-turn examination recommendation: LLMs use more comprehensive information from the free-turn setting;
(iii) oracle diagnosis: LLMs have access to all ground-truth examination evidences, representing the easiest setting.

\textbf{Treatment Planning}.

This setting evaluates LLMs’ ability to propose suitable treatment plans. Using the \textbf{MedR-Bench-Treatment}, case summaries—including diagnostic results—serve as input, with the real-world treatment plan as the reference. Unlike diagnosis, only the oracle setting is used, where LLMs are provided with all ground-truth patient data, for example, basic patient information, ancillary tests, and ground-truth diagnostic results. This reflects the challenges of treatment planning, which is sufficiently challenging, as suggested by our results.

\subsubsection{Evaluation Metrics}

We designed six metrics to objectively evaluate the performance of LLMs, focusing on both their reasoning processes and final outputs, as illustrated in Figure~\ref{fig:teaser}b. 
Notably, for DeepSeek-R1, it will have two potential reasoning parts, one presented in the formal part and the other presented in the default thinking part~(please refer to Method~\ref{sec:LLMBaselines} for more detailed explanations). By default, in figures, we report the former for fair comparison. In tables, we report reasoning metrics for both, recorded as “XX /xx,” where the former denotes the reasoning part in the formal answer part, and the latter denotes the marked thinking part. Below, we briefly introduce these metrics, with more detailed explanations provided in the \textbf{Method} section.

For \textbf{reasoning processes}, 
which are primarily expressed in free text and pose significant evaluation challenges~\cite{wu2024pmc,qiu2024towards,zhao2024ratescore,calamida2023radiology}, 
we developed a novel LLM-based system called the \textbf{Reasoning Evaluator}. This system decomposes, structures, and verifies reasoning steps. It identifies effective versus repetitive steps and evaluates their alignment with medical knowledge or guidelines by referencing online medical resources. If ground-truth reasoning references are available, the system further assesses whether all relevant steps have been included.
Please refer to the Method for more details.

Based on this pipeline, we define the following reasoning metrics:

\vspace{-0.3cm}
\begin{itemize}
\setlength\itemsep{3pt}
    \item \textbf{Efficiency}: evaluates whether each reasoning step contributes new insights toward the final answer rather than repeating or rephrasing previous results. It is calculated as the proportion of effective reasoning steps within the entire reasoning prediction.
    
    \item \textbf{Factuality}: assesses whether effective reasoning steps adhere to medical guidelines or factual knowledge. Similar to a `precision' score, it calculates the proportion of factually correct steps among all predicted effective reasoning steps.
    
    \item \textbf{Completeness}: measures how many reasoning steps explicitly marked in the raw case report are included in the generated content. Analogous to `recall', it computes the proportion of mentioned reasoning steps among all ground-truth steps. 
    While raw case reports may omit some steps, those included are considered essential reasoning evidence.    
    
\end{itemize}

\vspace{-0.2cm}
On \textbf{the final generation}, for example, recommended examinations, diagnosed diseases, and treatment plans, the following metrics are used:

\vspace{-0.3cm}
\begin{itemize}
\setlength\itemsep{3pt}
    \item \textbf{Accuracy}: evaluates whether the final answer (both diagnosis and treatment) explicitly matches the ground-truth provided in the raw case reports.
    
    \item \textbf{Precision \& Recall}: 
    used for examination recommendation, where LLMs generate a list of recommended examinations for a given patient case. These metrics are calculated by comparing the generated examination list with the ground-truth ancillary test list recorded in the case report.
\end{itemize}

\subsection{Results in Examination Recommendation}

This section presents the main evaluation results for examination recommendations, as illustrated in Figure~\ref{fig:teaser}c and Figure~\ref{fig:teaser}d. Detailed results for the recommended examinations are summarized in Extended Table~\ref{tab:assessment_results}, while the results for the reasoning processes are provided in Extended Table~\ref{tab:assessment_reasoning}.

\textbf{Analysis on Recommended Examinations.} 

In the \textbf{1-turn setting}, as shown in Extended Table~\ref{tab:assessment_results}, DeepSeek-R1 achieves the highest recall at 43.61\%, demonstrating its ability to identify the most relevant examinations. Gemini-2.0-FT follows closely with a recall of 43.12\%. 
Qwen-QwQ ranks in the middle, while OpenAI-o3-mini and Baichuan-M1 perform sub-optimally. 
However, their results remain comparable to those of their competitors.

For \textbf{precision}, Baichuan-M1 outperforms other models with a score of 41.78\%, 
indicating better alignment with medical scenarios and ability to recommend relevant examinations. In contrast, Qwen-QwQ and Gemini-2.0-FT score the lower precision at 24.43\% and 22.77\%, respectively, suggesting frequent recommendations of irrelevant examinations. 
This may be attributed to Qwen-QwQ's smaller parameter size and their optimization focus on mathematical domains rather than medical reasoning.

In the \textbf{free-turn setting}, 
where models are allowed unlimited queries, no significant improvements are observed in either precision or recall across all models. Missed examinations remain un-recovered, 
even with additional turns, and performance even declines slightly in some cases.
For example, OpenAI-o3-mini achieves a recall of 38.22\% in the free-turn setting, 
slightly lower than its 1-turn recall of 38.47\%. 
Similarly, DeepSeek-R1 drops from 43.61\% in the 1-turn setting to 40.67\% in the free-turn setting. 

A notable issue observed in the free-turn setting is that models frequently enter repetitive query loops, requesting examinations that have either already been provided or explicitly stated as unavailable in earlier turns. This inefficiency in handling multi-turn dialogues limits the utility of the free-turn setting and highlights the challenges current LLMs face in dynamically proposing new queries during extended interactions.

Finally, when analyzing performance on rare diseases~(Extended Table~\ref{tab:assessment_results}), we find that most models maintain comparable performance to that across common diseases. 

\textbf{Analysis on Reasoning Processes.}

At the reasoning level, we focus primarily on the \textbf{1-turn setting}, as the free-turn setting involves extended reasoning processes that grow with the number of turns. Notably, \textbf{completeness} cannot be calculated in this context because raw case reports rarely document the reasoning behind the selection of specific examinations.

As shown in Extended Table~\ref{tab:assessment_reasoning}, the results on \textbf{efficiency} reveal that DeepSeek-R1 achieves the highest score at 98.59\%, demonstrating its ability to produce concise and relevant reasoning steps. In contrast, Qwen-QwQ performs the worst, with an efficiency score of just 86.53\%. This may be attributed to its training objective of ``reflecting deeply''~\cite{qwq-32b-preview}, likely causing it to generate excessive attempts, ultimately reducing its efficiency.

For \textbf{factuality}, most LLMs perform well, achieving scores close to 95\%. Among them, Gemini-2.0-FT emerges as the most reliable model in examination recommendation, with a factuality score of 98.75\%. However, it is notable that none of the models achieve perfect factuality (100\%) in their reasoning processes, underscoring the need to carefully verify critical reasoning steps in real-world medical applications.

When analyzing reasoning on rare diseases (Extended Table~\ref{tab:assessment_reasoning}), 
we observe consistent trends with those for common diseases, suggesting the robustness of LLMs across common and rare cases.

\subsection{Results in Diagnostic Decision-Making}

This section presents the results for diagnostic decision-making, 
analyzing performance on both the final output and reasoning levels.

\textbf{Analysis on Disease Diagnosis.}

As shown in Figure~\ref{fig:teaser}c and Extended Table~\ref{tab:diagnostic_results}, 
we evaluate diagnostic performance across three settings: \textbf{1-turn}, \textbf{free-turn}, and \textbf{oracle}. Generally, we find DeepSeek-R1 achieves the best diagnosis accuracy among all these settings, with accuracy scores of 71.79\%, 76.18\%, and 89.76\%, respectively.

In the \textbf{1-turn setting}, DeepSeek-R1 achieves the highest diagnostic accuracy (71.79\%), demonstrating its ability to gather relevant information and produce accurate diagnoses. 
Gemini-2.0-FT follows with an accuracy of 68.55\%. 
These results highlight the correlation between active information collection and diagnostic precision. Baichuan-M1 ranks in the middle, while OpenAI-o3-mini and Qwen-QwQ perform less effectively, consistent with their results in examination recommendation.

In the \textbf{free-turn setting}, 
where models can iteratively query additional information, most models show improved diagnostic accuracy despite minimal changes in recall scores for examination recommendation. For instance, DeepSeek-R1 increases its accuracy from 71.79\% (1-turn) to 76.18\%, and OpenAI-o3-mini improves from 64.99\% to 67.19\%, likely stems from increased inference computation, where models re-analyze examination results across multiple turns. However, Baichuan-M1 experiences a significant drop in accuracy, possibly due to context length limitations associated with its smaller model size.

In the \textbf{oracle setting}, 
where all crucial diagnostic information is provided, all models achieve significantly higher accuracy. For example, {DeepSeek-R1} improves from 76.18\% in the free-turn setting to 89.76\%, followed by {Gemini-2.0-FT}.  OpenAI-o3-mini, Qwen-QwQ, and Baichuan-M1 also perform well, achieving accuracies above 83\%. These results emphasize the importance of identifying and recommending relevant examinations to support accurate diagnoses.

Overall, all models achieve over 80\% diagnostic accuracy in the oracle setting, indicating that when sufficient information is available, current LLMs can reliably perform diagnostic tasks. Performance on rare diseases is consistent with common ones, 
further demonstrating the robustness of these models in challenging scenarios.

\textbf{Analysis on Reasoning Processes.} 

In Figure~\ref{fig:teaser}d and Extended Table~\ref{tab:diagnostic_reasoning_all_diseases} , we show the results in diagnostic reasoning processes on all diseases. In Extended Table~\ref{tab:diagnostic_reasoning_rare_diseases}, we further show the results on rare diseases.

In the \textbf{1-turn diagnostic setting}, as shown Table~\ref{tab:diagnostic_reasoning_all_diseases}, in where reasoning builds on incomplete examinations, 
most models—except Qwen-QwQ—show a decline in factuality compared to the oracle setting. This suggests that missing examinations increases the likelihood of hallucinated reasoning. Interestingly, Qwen-QwQ performs relatively well in this context, achieving a completeness score of 66.94\%, despite its inefficiency, which prioritizes generating exhaustive reasoning at the expense of efficiency.

In the \textbf{oracle setting}, where all essential examination results are provided, Gemini-2.0-FT excels in factuality and completeness, achieving scores of 98.23\% and 83.28\%, respectively. DeepSeek-R1 demonstrates the highest efficiency at 97.17\%. 
In contrast, Qwen-QwQ performs poorly in efficiency (71.20\%) and factuality (84.02\%), 
due to its tendency for verbose reasoning. However, this verbosity enables it to retrieve more ground-truth reasoning evidence, resulting in the highest completeness score among open-source models~(79.97\%). Notably, for rare diseases as shown in Table~\ref{tab:diagnostic_reasoning_rare_diseases}, the performance remains consistent, 
and the factuality of most LLMs does not decline.

\subsection{Results in Treatment Planning}

This section presents the results of treatment planning. The overall findings are illustrated in Figure~\ref{fig:teaser}c (final generation) and Figure~\ref{fig:teaser}d (reasoning processes), with detailed results provided in Extended Table~\ref{tab:treatment}.

\textbf{Analysis on Treatment Plans.} 

In treatment planning, we observe that the precision of recommended treatment plans is significantly lower than the accuracy of diagnostic outputs. Among the models, 
Baichuan-M1 and DeepSeek-R1 achieves the highest precision at 30.65\% and 30.51\%, respectively. These results underline the increased complexity of treatment planning compared to diagnosis, emphasizing the need for further development of LLMs.

Unlike diagnosis, where rare cases do not impact performance, treatment planning shows a notable decline in precision for rare diseases across general models. 
For instance, OpenAI-o3-mini drops from 27.03\% to 23.17\%, and DeepSeek-R1 decreases from 30.51\% to 27.27\%. This highlights a persistent gap in therapeutic knowledge for rare conditions. In contrast, Baichuan-M1 maintains stable performance, with precision only slightly decreasing from 30.65\% to 30.30\%, demonstrating the effectiveness of its medical knowledge enhancement.

\textbf{Analysis on Reasoning Processes.} 

As shown in Extended Table~\ref{tab:treatment}, reasoning quality in treatment planning is generally strong. Most models achieve high factuality scores exceeding 95\%, indicating alignment with medical guidelines. DeepSeek-R1 achieves the highest efficiency (95.25\%), while Gemini-2.0-FT leads in factuality (96.96\%), followed by OpenAI-o3-mini (96.77\%). Qwen-QwQ, consistent with its performance in other stages, exhibits the lowest efficiency (84.76\%) but achieves the highest completeness (77.66\%), reflecting its tendency to generate extensive reasoning. 
Interestingly, for rare cases, the reasoning performance does not change significantly, as models maintain high factuality and completeness scores. However, in final treatment planning decisions, the models generally show low accuracy.


\section{Discussion}

In this study, we evaluate the latest reasoning-enhanced large language models (LLMs) in the medical domain, focusing on both final outputs and the underlying reasoning processes. Unlike previous work on medical LLMs evaluation~\cite{jin-etal-2019-pubmedqa, jin2021disease, pal2022medmcqa, wu2025towards, singhal2023large, singhal2025toward, wu2024pmc, qiu2024towards, xie2024preliminary, hager2024evaluation, lamparth2025moving}, our approach places greater emphasis on quantifying the quality of reasoning. The key contributions of this study are as follows:

\textbf{A diverse evaluation dataset on real-world patient cases with reasoning references.} We introduce \textbf{MedR-Bench}, a dataset of 1,453 structured patient cases derived from published case reports. It spans 13 medical body systems and 10 disorder specialties, covering both common and rare diseases for diagnosis and treatment planning. Unlike existing multiple-choice datasets, MedR-Bench closely mirrors real-world medical practice. Furthermore, each case is enriched with reasoning evidence extracted from high-quality case reports, enabling a rigorous evaluation of reasoning processes.

\textbf{A versatile evaluation framework covering three critical patient stages.} 
Our benchmark assesses LLM performance across three key stages of patient care: examination recommendation, diagnostic decision-making, and treatment planning. 
This framework replicates a typical clinical workflow, 
providing insights into areas where LLMs perform well and identifying gaps in their ability to support clinical decision-making.

\textbf{A set of objective metrics from multiple perspectives.} 
We adopt a multi-dimensional set of metrics to assess LLM performance. 
Beyond evaluating the accuracy of final outputs, we introduce the \textbf{Reasoning Evaluator}, a system designed to quantitatively measure the quality of free-text reasoning. Using an automated verification mechanism, this system ensures that reasoning is supported by authoritative medical evidence or aligns with reference ground-truth reasoning.

The following findings summarize the performance of LLMs on MedR-Bench:

\textbf{LLMs demonstrate strong diagnostic performance with sufficient examinations.} 
State-of-the-art reasoning LLMs demonstrate strong diagnostic capabilities when presented with sufficient, well-structured information~(Table~\ref{tab:diagnostic_results}). 
These models excel at synthesizing medical examination results from different specialists to produce clear and accurate diagnoses. While occasional mistakes occur, the overall results are encouraging and highlight the potential for integrating LLMs into clinical workflows. This represents a promising step toward their integration into medical practice.

\textbf{Examination recommendation and treatment planning remain challenging.} 
Despite their diagnostic success, LLMs struggle with recommending additional examinations to gather necessary diagnostic clues (Table~\ref{tab:assessment_results}). This limitation is critical, as such recommendations are central to accurate medical decision-making. Similarly, treatment planning poses significant challenges, with performance in this area lagging notably. This shortfall is likely due to the fact that oracle diagnosis setting closely resembles multiple-choice medical question-answering datasets~\cite{pal2022medmcqa,jin2021disease}, which have been widely optimized. 
This suggests that while LLMs have mastered basic medical knowledge and can synthesize examination results, they are not yet aligned with the dynamic, real-world hospital environment. These gaps underscore the need for human oversight in clinical applications and highlight key areas for future improvement.

\textbf{Reasoning capabilities of LLMs in medicine remain inadequate.} 
Our benchmark evaluates reasoning quality through metrics such as efficiency, factuality, and completeness. While most models achieve high efficiency (over 90\%, except for Qwen-QwQ), indicating that their reasoning steps contribute meaningfully to decisions, factuality scores reveal occasional errors. Such mistakes, though tolerable in general contexts, pose risks in clinical settings where over-reliance on LLM outputs could lead to harm. Completeness is particularly concerning, with scores between 70\% and 80\%, reflecting frequent omissions of critical reasoning steps essential for clinical decision-making. Overall, the reasoning capabilities of current LLMs are barely satisfactory and require substantial improvement to meet the demands of clinical reliability and accuracy.

\textbf{LLMs maintain robust performance on rare diseases, despite challenges.} 
MedR-Bench includes cases involving rare diseases, which are inherently more difficult. While performance in treatment planning for these cases is weaker,
the decline is mild, and LLMs maintain consistent accuracy in other tasks. This robustness suggests that current LLMs possess a strong foundational understanding of medical knowledge, even for rare disease domains, underscoring their potential applicability in diverse clinical scenarios.

\textbf{The gap between open-source and closed-source LLMs is narrowing.} 
Encouragingly, the latest open-source models, such as DeepSeek-R1, are approaching the performance of closed-source LLMs in medical tasks. Open-source models offer significant advantages for clinical applications, including local deployment to safeguard patient privacy and mitigate risks of data leakage. Their accessibility also reduces reliance on proprietary systems, fostering broader adoption of LLM-driven solutions in medicine while avoiding monopolization of medical resources.

\textbf{Limitations.} This work has several limitations. 
{\em First}, while we ensured the evaluation cases were derived from recent case reports, we cannot fully guarantee that all cases were excluded from the training data of the evaluated models, as some LLMs do not disclose their training data cutoff dates. {\em Second}, patient cases in MedR-Bench were automatically converted by LLMs, and while supported by case reports, errors could have been introduced during this process. {\em Third}, the metrics designed to evaluate LLM performance, while objective and scalable, cannot fully replace human verification. Human review, though resource-intensive, remains essential for validating clinical accuracy.

To address these limitations, we have released all code, evaluation cases, and model responses for the community to access and refine. We encourage clinicians to engage in reviewing and validating LLM-generated responses to further advance research in this domain.

\clearpage
\section{Methods}

This section describes the development of MedR-Bench, 
including the data curation pipeline, the three-stage evaluation framework, 
and the implementation of evaluation metrics via the Reasoning Evaluator. 
All text prompts used are provided in the supplementary materials and referenced as \textit{Prompt}~X, where \textit{X} denotes the corresponding prompt number.

\subsection{Data Curation}

\begin{figure}[t!]
    \centering
    \includegraphics[width=\linewidth]{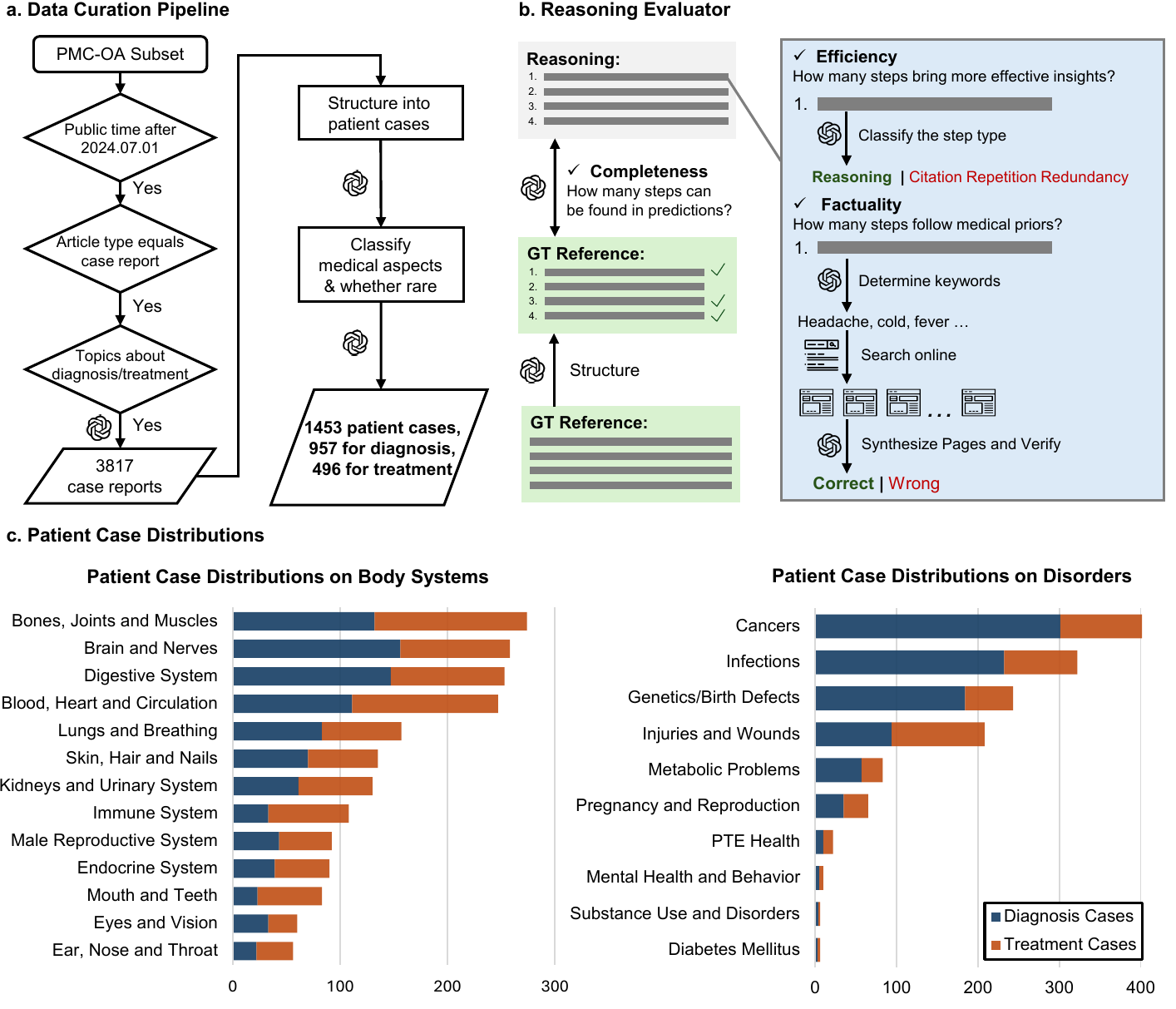}
    \caption{\textbf{Overview of our data curation pipeline, Reasoning Evaluator, and final patient case distributions.} \textbf{a} illustrates our data curation pipeline using a flowchart. We start with the original case reports from the PMC-OA subset, then filter and reorganize them into structured patient cases for testing. \textbf{b} depicts our Reasoning Evaluator to quantitatively measure reasoning quality from efficiency, factuality, and completeness aspects. External search engines are employed to assist the agent in more accurately evaluating the correctness of the provided reasoning steps. \textbf{c} This figure presents the distribution of patient cases across different medical aspects. }
    \label{fig:method}
\end{figure}

As illustrated in Figure~\ref{fig:method}a, 
case reports were collected from the PubMed Central Open Access (PMC-OA) Subset~\cite{pmc_open_access_subset}, focusing on articles labeled as `case reports'. To minimize potential data leakage, we excluded papers published before July 2024, aligning with the training data cut-off date of OpenAI-o3-mini and Gemini-2.0-FT. While other models did not disclose their cutoff dates, their release timeline (near January 2025) and comparable performance suggest this cut-off date is adequate for analysis. This filtering yielded 3,817 raw case reports.

To ensure relevance, case reports unrelated to diagnosis or treatment, 
such as those focused on medical education, were excluded using GPT-4o~\cite{openai_gpt4o} (gpt-4o-2024-11-20) with \textit{Prompt}~\ref{prompt:prompt1_classify_casereport_category}.
Relevant reports were reformatted into structured patient cases using GPT-4o. Diagnosis-related cases included sections on ``differential diagnosis processes'' and ``final diagnosis explanations''~(\textit{Prompt}~\ref{prompt:prompt2_generate_diagnosis_case}), 
while treatment-related cases included ``treatment objectives'' and ``comprehensive rationale''~(\textit{Prompt}~\ref{prompt:prompt3_generate_treatmentplanning_case}).

\noindent \textbf{Patient Case Classification.} 

To ensure comprehensive coverage of patient cases in our evaluation dataset, 
each case is classified based on medical aspects and its relevance to rare diseases. For medical aspects, we adopt the ``Body System'' and ``Disorders and Conditions'' taxonomies from MedlinePlus~\cite{medlineplus}, as outlined on their ``Health Topics'' page. Cases that do not fit into any predefined category are classified as ``others''. 
Using \textit{Prompt}~\ref{prompt:prompt4_body_system_classification}, GPT-4o categorizes cases into body system classes based on the primarily affected body part, while \textit{Prompt}~\ref{prompt:prompt4_disorder_system_classification} assigns cases to disorder categories based on the associated diseases.

To identify rare disease cases, we use the Rare Disease Ontology (ORDO\footnote{\url{http://www.ebi.ac.uk/ols4/ontologies/ordo}}) provided by Orphanet~\cite{weinreich2008orphanet}. First, \texttt{Scispacy}~\cite{neumann-etal-2019-scispacy} is employed to extract all associated UMLS~\cite{bodenreider2004unified} Concept Unique Identifiers (CUIs) from the patient case. If any CUIs match those listed in ORDO, the case undergoes further verification using GPT-4o at the free-text level with {\em Prompt}~\ref{prompt:prompt5_check_if_rare_disease} to confirm explicit mention of rare diseases. Cases passing both steps are classified as rare disease-related; otherwise, they are labeled as unrelated to rare diseases.

Consequently, all patient cases are categorized into three dimensions: 
``Body System'', ``Disorders and Conditions''~(abbreviated as ``Disorder''), 
and rare disease relevance. In total, \textbf{1,453} real-world patient cases are included in \textbf{MedR-Bench}, comprising \textbf{957} diagnosis cases and \textbf{496} treatment cases. Among these, \textbf{491} diagnosis cases and \textbf{165} treatment cases are related to rare diseases. 
The distribution of cases across medical aspects is shown in Figure~\ref{fig:method}a. Detailed patient cases, along with reference category labels, are provided in Supplementary~\ref{case_study}.

\subsection{Evaluation Framework}

In this section, we introduce the implementation details of our evaluation framework. Three critical patient stages are considered: examination recommendation, diagnostic decision-making, and treatment planning.

\noindent \textbf{Examination Recommendation}. 

In this stage, inspired by prior works~\cite{hager2024evaluation, johri2025evaluation, liao2024automatic}, we evaluate the ability of LLMs to dynamically interact with patients and actively recommend necessary examinations for clinical decision-making. To achieve this, we build a patient agent using GPT-4o with {\em Prompt}~\ref{prompt:prompt6_dynamic_diagnosis_gpt4o_giving_info}, where \{case\} represents a specific patient case from \textbf{MedR-Bench}. 
The patient agent is designed to role-play as a virtual patient, enabling dynamic interactions with LLMs and providing responses to their queries.

During evaluation, clinical LLMs are presented with a patient case summary, excluding details about ancillary tests, and tasked with interacting with the patient agent to gather the required information for an accurate diagnosis. The interaction follows one of two protocols: \textbf{1-turn examination recommendation} or \textbf{free-turn examination recommendation}. 
In each turn, LLMs may request additional examinations, such as imaging or lab tests, simulating real-world clinical workflows. If the requested examination is not available in the patient case, the patient agent will respond with: ``There is no relevant ancillary test information available for this request.''

Under the 1-turn protocol, LLMs are prompted to request essential additional information based on the patient case using {\em Prompt}~\ref{prompt:prompt7_askinfo_dynamic_diagnose}. 
In the free-turn protocol, LLMs are first prompted with {\em Prompt}~\ref{prompt:prompt7_first_turn_free_turn_dynamic_screening} to input the patient case summary. For subsequent turns, they are prompted using {\em Prompt}~\ref{prompt:prompt8_subsequent_turns_free_turn_dynamic_screening} to decide whether the available information is sufficient to make a clear diagnosis.

\noindent \textbf{Diagnostic Decision-making}.

In this stage, we assess the LLM's diagnostic capabilities across different settings, ordered by increasing critical information availability:
(i) \textbf{diagnosis after 1-turn examination recommendation:} The LLMs are prompted to provide a final diagnosis by integrating the basic patient case information with the additional details obtained during the 1-turn examination recommendation stage, using \textit{Prompt}~\ref{prompt:prompt8_make_final_diagnosis_dynamic_diagnose}.
(ii) \textbf{diagnosis after free-turn examination recommendation:} In this scenario, the LLMs diagnose based on examination information gathered during free-turn interactions, where they determine that the available information is sufficient. To prevent infinite loops, the maximum number of turns is capped at five. If this limit is reached, the LLM is required to make a diagnosis based on the information collected up to that point.
(iii) \textbf{oracle diagnosis:} In this setting, the LLMs are provided with the full ground-truth patient information, including all auxiliary tests, and are prompted to make a diagnosis using \textit{Prompt}~\ref{prompt:prompt9_accurare_diagnose}.

\noindent \textbf{Treatment Planning}.

In this stage, we provide the LLMs with the complete patient information, including the final diagnosis result, to generate recommendations for the preferred treatment plans using \textit{Prompt}~\ref{prompt:prompt10_treament_plan}. 
Specifically, for each patient case in MedR-Bench, the complete case summary is provided as input~(oracle evaluation), and the LLMs are instructed to perform treatment planning.

\subsection{Evaluation Metrics}
In this section, we provide a detailed explanation of the implementation of various evaluation metrics.

To begin with, \textbf{at the reasoning level}, we introduce \textbf{Reasoning Evaluator}, an agentic system powered by GPT-4o, designed to objectively assess the quality of free-text reasoning, as shown in Figure~\ref{fig:method}b. Formally, let the predicted reasoning process be denoted as $\mathcal{\hat{R}} = \{\hat{r}_1, \hat{r}_2, \cdots, \hat{r}_N\}$, where each $\hat{r}_i$ represents a reasoning step generated by the original assessed LLMs.
The system begins by evaluating the effectiveness of each reasoning step, classifying each step into one of four categories: \{citation, repetition, redundancy, reasoning\}.
\begin{itemize}
\setlength\itemsep{3pt}
    \item \textbf{Citation} refers to steps that solely restate or cite information directly from the input.
    \item \textbf{Repetition} refers to steps that merely restate conclusions already made in earlier reasoning steps.
    \item \textbf{Redundancy} denotes steps that do not contribute meaningfully to the final decision and are irrelevant to the reasoning process.
    \item \textbf{Reasoning} refers to steps that provide additional insights and contribute to the final decision.
\end{itemize}
Only steps classified as reasoning are considered effective. Formally, this classification can be formulated as $e_i = \mathcal{A}(\hat{r}_i \mid P_e)$, where $e_i \in \{0,1\}$ indicates whether a given step is effective, and $P_e$ represents the prompt (\textit{Prompt}~\ref{prompt:prompt12_effi_classification}) used to instruct GPT-4o. 

Afterward, the agentic system evaluates the factuality of each effective reasoning step by verifying its consistency with external medical knowledge or established guidelines. Specifically, the system first generates a series of search keywords for each effective reasoning step, which is formulated as:
\begin{equation}
    \mathcal{K} = \mathcal{A}(\hat{r}_i \mid P_k),\quad\text{if } e_i = 1,
\end{equation}
where $\mathcal{K}$ denotes the search keyword set and $P_k$ represents the related prompts (\textit{Prompt}~\ref{prompt:prompt13_get_keywords}). By interacting with external search engine tools, including Google\footnote{www.google.com}, Bing\footnote{www.bing.com}, or DuckDuckGo\footnote{www.duckduckgo.com}, we can retrieve the Top-3 recommended online pages. The system will then summarize their information as the environment response, formulated as $\text{Response} = \mathcal{A}(\text{Search}(\mathcal{K}) \mid P_s)$, where $\text{Search}(\cdot)$ represents the search APIs and $P_s$ is the prompt used for summarization. 
Finally, the agentic system determines the correctness of each step based on the summarized response:
\begin{equation}
    c_i = \begin{cases} 
    0, & \text{if } e_i = 0, \\
    \mathcal{A}(\hat{r}_i \mid \text{Response}, P_c), & \text{if } e_i = 1.
\end{cases}
\end{equation}
Similarly, here, $P_c$ is the prompt (\textit{Prompt}~\ref{prompt:prompt14_check_factuality}) used to evaluate whether the model output is consistent with the searched factual information or contradicts it.

Next, if the ground truth reasoning evidence $\mathcal{R}$ is provided, the agentic system will be employed to compare it against the prediction. It evaluates how many steps of the ground truth reasoning evidence can be found within the prediction $\hat{\mathcal{R}}$. 
We first decompose $\mathcal{R}$ into multiple steps as $\{{r}_1, {r}_2, \dots, {r}_M\} = \mathcal{A}({\mathcal{R}} \mid P_d)$ using \textit{Prompt}~\ref{prompt:prompt11_reformat_unstructured_ground_truth_rationale}. Then, we prompt the system with \textit{Prompt}~\ref{prompt:prompt15_check_hit} to determine whether each step can be found in the prediction using $P_f$:
\begin{equation}
    f_i = \mathcal{A}({r}_i, \hat{\mathcal{R}} \mid P_f).
\end{equation}

Based on the results obtained from the agentic reasoning judgment process, the following reasoning-related metrics can be derived:
\begin{itemize}
\setlength\itemsep{3pt}
    \item \textbf{Efficiency}: This metric evaluates the extent to which reasoning steps contribute additional insights toward the final answer, rather than merely repeating previous results or invoking irrelevant reasoning content. The efficiency score is defined as:
    \begin{equation}
        \text{Efficiency} = \frac{1}{N} \sum_{i=1}^{N} e_i,
    \end{equation}
    
    \item \textbf{Factuality:} In this metric, we focus on evaluating the factual accuracy of reasoning steps. This can be analogous to Precision scores. Based on the results of the \textbf{Reasoning Evaluator}, we calculate the proportion of steps that adhere to established medical knowledge or guidelines among all effective steps:
    \begin{equation}
        \text{Factuality} = \frac{\sum^{N}_{i=1} c_i}{\sum^{N}_{i=1} e_i},
    \end{equation}
    
    \item \textbf{Completeness}: This metric assesses the extent to which reasoning steps outlined in raw case reports are reflected in the generated content. It is analogous to Recall scores and is calculated as:
    \begin{equation}
        \text{Completeness} = \frac{1}{M}\sum^{M}_{i=1} f_i .
    \end{equation}
    
\end{itemize}

To further evaluate the reliability of these metrics, we conducted a manual verification of the key classification steps, \emph{i.e.}, effectiveness classification, factuality judgment, and completeness assessment. For each component, we sample 100 cases and assigned them to four independent evaluators to verify the accuracy of the results predicted by the agentic system. Our system achieves accuracies of 98\%, 99\%, and 90\%, respectively. These results demonstrate the validity of the final metrics.

Additionally, \textbf{at the final generation level}, 
{\em e.g.}, examination recommendation, disease diagnosis, treatment planning, 
we adopt several classical metrics to quantify performance:
\begin{itemize}
\setlength\itemsep{3pt}
    \item \textbf{Accuracy}: This metric is a binary metric. It directly compares whether the final answer clearly matches the ground truth provided in the raw case reports. Since medical terminologies often have synonyms, we utilize GPT-4o to verify whether the predicted results are equivalent to the ground truth. For accurate diagnosis, we employ the prompt described in \textit{Prompt}~\ref{prompt:prompt9_diagnosis_accuracy_judgement}. In contrast, treatment planning is more complex than accurate diagnosis, as even the same disease can have multiple treatment pathways. To address this complexity, we first extract keywords from patient cases using \textit{Prompt}~\ref{prompt:prompt13_get_keywords}. Subsequently, we use a search engine to gather relevant information and make a judgment based on both the retrieved information and the ground-truth treatment plan, as described in \textit{Prompt}~\ref{prompt:prompt10_treament_planing_accuracy_judgement}.

    \item \textbf{Precision and Recall}: These metrics are employed in the context of examination recommendation. They compare the recommended examination list generated by the LLM against the ground-truth practical list using list-wise precision and recall scores. Since the LLM's queries are presented in free-text format, we first utilize GPT-4o to summarize and reorganize them into a structured list using \textit{Prompt}~\ref{prompt:prompt17_reformat_addtional_infomation_required}. Subsequently, we use \textit{Prompt}~\ref{prompt:prompt15_check_hit} to evaluate the hit rate.

\end{itemize}

Similarly, we conducted manual checks on 100 sampled cases to verify whether GPT-4o could accurately evaluate diagnostic and treatment planning predictions compared to the ground truth. In the diagnostic task, GPT-4o achieved an accuracy rate of 96\%. For the treatment planning task, which is inherently more complex due to its variability, GPT-4o achieved an accuracy rate of 84\%.

\subsection{LLM Baselines}
\label{sec:LLMBaselines}
In our \textbf{MedR-Bench}, we evaluated five mainstream reasoning LLM series:
\vspace{-0.3cm}
\begin{itemize}
\setlength\itemsep{3pt}
    \item \textbf{OpenAI-o3-mini}~\cite{openai_o3_mini}: The \texttt{o3-mini} is the latest LLM developed by OpenAI and is widely regarded as the most powerful LLM currently available. Compared to OpenAI's previous model, GPT-4o, its most notable feature is its enhanced reasoning ability, or, in other words, its capability to ``think'' before answering. We evaluated the model version \texttt{o3-mini-2025-01-31} using the official API.
    

    \item \textbf{Gemini-2.0-Flash-Thinking~(FT)}~\cite{team2023gemini}: The \texttt{Gemini-2.0-Flash-Thinking} is an experimental ``thinking'' LLM developed by Google. It exhibits stronger reasoning capabilities in its responses compared to its predecessor, the Gemini 2.0 Flash Experimental model. This model is characterized by its explicit ``thinking process'' prior to generating answers. We evaluated the model version \texttt{gemini-2.0-flash-thinking-exp-01-21} using the official API.
    
    \item \textbf{DeepSeek-R1}~\cite{guo2025deepseek}: DeepSeek-R1 is a 671B-parameter LLM developed by the DeepSeek company. It is an open-source model and is regarded as achieving performance comparable to OpenAI's o1. Similar to o1, it is a reasoning LLM, capable of producing explicit ``thinking'' outputs. In our evaluation, we use the model weights from Hugging Face, \texttt{deepseek-ai/DeepSeek-R1}\footnote{\url{https://huggingface.co/deepseek-ai/DeepSeek-R1}}, and deploy it locally.
    \item \textbf{Qwen-QWQ}~\cite{qwq-32b-preview}:Qwen-QwQ is a 32B-parameter experimental research model developed by the Qwen Team. Similar to OpenAI-o1 and DeepSeek-R1, it is also focused on advancing LLM reasoning capabilities. We use the model weights from \texttt{Qwen/QwQ-32B-Preview}\footnote{\url{https://huggingface.co/Qwen/QwQ-32B-Preview}} and deploy it locally for evaluation.
    
    \item \textbf{Baichuan-M1}~\cite{wang2025baichuan}: Baichuan-M1 is a 14B-parameter medical-specific LLM developed by the Baichuan company. Unlike the previously mentioned models, which are designed for general domains, Baichuan-M1 is the latest specialized medical LLM. We use the model \texttt{baichuan-inc/Baichuan-M1-14B-Instruct}\footnote{\url{https://huggingface.co/baichuan-inc/Baichuan-M1-14B-Instruct}} deployed locally for evaluation.
\end{itemize}

\textbf{Notably}, during evaluation, there are two ways to obtain a model's reasoning responses. One approach is to use the model's default marked ``thinking parts.'' For instance, in the case of \textbf{DeepSeek-R1}, its responses always consist of two distinct parts: a thinking part and a formal answer part, separated by the special tokens ``<think>'' and ``</think>.'' The output format of \textbf{OpenAI-o3-mini} follows the same structure. While it seems natural to consider the thinking part as reasoning, \textbf{OpenAI-o3-mini} omits this by default and other models, such as \textbf{Qwen-QWQ}, \textbf{Baichuan-M1}, and \textbf{Gemini-2.0-Flash-Thinking~(FT)}, do not make such a distinction between the reasoning and answer parts. Thus, to standardize reasoning evaluation across all models, we employ the second approach to obtain reasoning: prompting them with ``summarize the reasoning step-by-step'' to explicitly instruct them to generate reasoning responses. 
For \textbf{DeepSeek-R1}, this approach results in two potential reasoning outputs: the reasoning response generated within the formal answer part and an additional thinking part marked by the special tokens. By default, in figures, we report the former for fair comparison. In tables, we report reasoning metrics for both, recorded as ``XX\scalebox{0.7}{ /xx},'' where the former denotes the reasoning part in the formal answer part, and the latter denotes the marked thinking part.

\clearpage
\bibliographystyle{unsrt}
\bibliography{references} 

\begin{thebibliography}{10}

\bibitem{jaech2024openai}
Aaron Jaech, Adam Kalai, Adam Lerer, Adam Richardson, Ahmed El-Kishky, Aiden Low, Alec Helyar, Aleksander Madry, Alex Beutel, Alex Carney, et~al.
\newblock Openai o1 system card.
\newblock {\em arXiv preprint arXiv:2412.16720}, 2024.

\bibitem{guo2025deepseek}
Daya Guo, Dejian Yang, Haowei Zhang, Junxiao Song, Ruoyu Zhang, Runxin Xu, Qihao Zhu, Shirong Ma, Peiyi Wang, Xiao Bi, et~al.
\newblock Deepseek-r1: Incentivizing reasoning capability in llms via reinforcement learning.
\newblock {\em arXiv preprint arXiv:2501.12948}, 2025.

\bibitem{zhong2024evaluation}
Tianyang Zhong, Zhengliang Liu, Yi~Pan, Yutong Zhang, Yifan Zhou, Shizhe Liang, Zihao Wu, Yanjun Lyu, Peng Shu, Xiaowei Yu, et~al.
\newblock Evaluation of openai o1: Opportunities and challenges of agi.
\newblock {\em arXiv preprint arXiv:2409.18486}, 2024.

\bibitem{phan2025humanity}
Long Phan, Alice Gatti, Ziwen Han, Nathaniel Li, Josephina Hu, Hugh Zhang, Sean Shi, Michael Choi, Anish Agrawal, Arnav Chopra, et~al.
\newblock Humanity's last exam.
\newblock {\em arXiv preprint arXiv:2501.14249}, 2025.

\bibitem{jin-etal-2019-pubmedqa}
Qiao Jin, Bhuwan Dhingra, Zhengping Liu, William Cohen, and Xinghua Lu.
\newblock {P}ub{M}ed{QA}: A dataset for biomedical research question answering.
\newblock In Kentaro Inui, Jing Jiang, Vincent Ng, and Xiaojun Wan, editors, {\em Proceedings of the 2019 Conference on Empirical Methods in Natural Language Processing and the 9th International Joint Conference on Natural Language Processing (EMNLP-IJCNLP)}, pages 2567--2577, Hong Kong, China, November 2019. Association for Computational Linguistics.

\bibitem{jin2021disease}
Di~Jin, Eileen Pan, Nassim Oufattole, Wei-Hung Weng, Hanyi Fang, and Peter Szolovits.
\newblock What disease does this patient have? a large-scale open domain question answering dataset from medical exams.
\newblock {\em Applied Sciences}, 11(14):6421, 2021.

\bibitem{pal2022medmcqa}
Ankit Pal, Logesh~Kumar Umapathi, and Malaikannan Sankarasubbu.
\newblock Medmcqa: A large-scale multi-subject multi-choice dataset for medical domain question answering.
\newblock In {\em Conference on health, inference, and learning}, pages 248--260. PMLR, 2022.

\bibitem{wu2025towards}
Chaoyi Wu, Pengcheng Qiu, Jinxin Liu, Hongfei Gu, Na~Li, Ya~Zhang, Yanfeng Wang, and Weidi Xie.
\newblock Towards evaluating and building versatile large language models for medicine.
\newblock {\em npj Digital Medicine}, 8(1):58, 2025.

\bibitem{singhal2023large}
Karan Singhal, Shekoofeh Azizi, Tao Tu, S~Sara Mahdavi, Jason Wei, Hyung~Won Chung, Nathan Scales, Ajay Tanwani, Heather Cole-Lewis, Stephen Pfohl, et~al.
\newblock Large language models encode clinical knowledge.
\newblock {\em Nature}, 620(7972):172--180, 2023.

\bibitem{singhal2025toward}
Karan Singhal, Tao Tu, Juraj Gottweis, Rory Sayres, Ellery Wulczyn, Mohamed Amin, Le~Hou, Kevin Clark, Stephen~R Pfohl, Heather Cole-Lewis, et~al.
\newblock Toward expert-level medical question answering with large language models.
\newblock {\em Nature Medicine}, pages 1--8, 2025.

\bibitem{wu2024pmc}
Chaoyi Wu, Weixiong Lin, Xiaoman Zhang, Ya~Zhang, Weidi Xie, and Yanfeng Wang.
\newblock Pmc-llama: toward building open-source language models for medicine.
\newblock {\em Journal of the American Medical Informatics Association}, page ocae045, 2024.

\bibitem{qiu2024towards}
Pengcheng Qiu, Chaoyi Wu, Xiaoman Zhang, Weixiong Lin, Haicheng Wang, Ya~Zhang, Yanfeng Wang, and Weidi Xie.
\newblock Towards building multilingual language model for medicine.
\newblock {\em Nature Communications}, 15(1):8384, 2024.

\bibitem{xie2024preliminary}
Yunfei Xie, Juncheng Wu, Haoqin Tu, Siwei Yang, Bingchen Zhao, Yongshuo Zong, Qiao Jin, Cihang Xie, and Yuyin Zhou.
\newblock A preliminary study of o1 in medicine: Are we closer to an ai doctor?
\newblock {\em arXiv preprint arXiv:2409.15277}, 2024.

\bibitem{hager2024evaluation}
Paul Hager, Friederike Jungmann, Robbie Holland, Kunal Bhagat, Inga Hubrecht, Manuel Knauer, Jakob Vielhauer, Marcus Makowski, Rickmer Braren, Georgios Kaissis, et~al.
\newblock Evaluation and mitigation of the limitations of large language models in clinical decision-making.
\newblock {\em Nature Medicine}, pages 1--10, 2024.

\bibitem{lamparth2025moving}
Max Lamparth, Declan Grabb, Amy Franks, Scott Gershan, Kaitlyn~N Kunstman, Aaron Lulla, Monika~Drummond Roots, Manu Sharma, Aryan Shrivastava, Nina Vasan, et~al.
\newblock Moving beyond medical exam questions: A clinician-annotated dataset of real-world tasks and ambiguity in mental healthcare.
\newblock {\em arXiv preprint arXiv:2502.16051}, 2025.

\bibitem{zhao2022pmc}
Zhengyun Zhao, Qiao Jin, Fangyuan Chen, Tuorui Peng, and Sheng Yu.
\newblock A large-scale dataset of patient summaries for retrieval-based clinical decision support systems.
\newblock {\em Scientific data}, 10(1):909, 2023.

\bibitem{wu2023towards}
Chaoyi Wu, Xiaoman Zhang, Ya~Zhang, Yanfeng Wang, and Weidi Xie.
\newblock Towards generalist foundation model for radiology.
\newblock {\em arXiv preprint arXiv:2308.02463}, 2023.

\bibitem{pmc_open_access_subset}
{National Library of Medicine}.
\newblock Pmc open access subset [internet].
\newblock \url{https://pmc.ncbi.nlm.nih.gov/tools/openftlist/}, 2003.
\newblock Bethesda (MD).

\bibitem{johri2025evaluation}
Shreya Johri, Jaehwan Jeong, Benjamin~A Tran, Daniel~I Schlessinger, Shannon Wongvibulsin, Leandra~A Barnes, Hong-Yu Zhou, Zhuo~Ran Cai, Eliezer~M Van~Allen, David Kim, et~al.
\newblock An evaluation framework for clinical use of large language models in patient interaction tasks.
\newblock {\em Nature Medicine}, pages 1--10, 2025.

\bibitem{liao2024automatic}
Yusheng Liao, Yutong Meng, Yuhao Wang, Hongcheng Liu, Yanfeng Wang, and Yu~Wang.
\newblock Automatic interactive evaluation for large language models with state aware patient simulator.
\newblock {\em arXiv preprint arXiv:2403.08495}, 2024.

\bibitem{zhao2024ratescore}
Weike Zhao, Chaoyi Wu, Xiaoman Zhang, Ya~Zhang, Yanfeng Wang, and Weidi Xie.
\newblock {R}a{TES}core: A metric for radiology report generation.
\newblock In Yaser Al-Onaizan, Mohit Bansal, and Yun-Nung Chen, editors, {\em Proceedings of the 2024 Conference on Empirical Methods in Natural Language Processing}, pages 15004--15019, Miami, Florida, USA, November 2024. Association for Computational Linguistics.

\bibitem{calamida2023radiology}
Amos Calamida, Farhad Nooralahzadeh, Morteza Rohanian, Koji Fujimoto, Mizuho Nishio, and Michael Krauthammer.
\newblock Radiology-aware model-based evaluation metric for report generation.
\newblock {\em arXiv preprint arXiv:2311.16764}, 2023.

\bibitem{qwq-32b-preview}
Qwen Team.
\newblock Qwq: Reflect deeply on the boundaries of the unknown, November 2024.
\newblock Accessed: 2025-02-27.

\bibitem{openai_gpt4o}
OpenAI.
\newblock Hello gpt-4o, 2025.
\newblock Accessed: 2025-02-27.

\bibitem{medlineplus}
{National Library of Medicine (US)}.
\newblock Medlineplus [internet], 2020.
\newblock [updated Jun 24; cited 2020 Jul 1].

\bibitem{weinreich2008orphanet}
Steffanie~S Weinreich, R~Mangon, JJ~Sikkens, ME~En Teeuw, and MC~Cornel.
\newblock Orphanet: a european database for rare diseases.
\newblock {\em Nederlands tijdschrift voor geneeskunde}, 152(9):518--519, 2008.

\bibitem{neumann-etal-2019-scispacy}
Mark Neumann, Daniel King, Iz~Beltagy, and Waleed Ammar.
\newblock {S}cispa{C}y: {F}ast and {R}obust {M}odels for {B}iomedical {N}atural {L}anguage {P}rocessing.
\newblock In {\em Proceedings of the 18th BioNLP Workshop and Shared Task}, pages 319--327, Florence, Italy, August 2019. Association for Computational Linguistics.

\bibitem{bodenreider2004unified}
Olivier Bodenreider.
\newblock The unified medical language system (umls): integrating biomedical terminology.
\newblock {\em Nucleic acids research}, 32(suppl\_1):D267--D270, 2004.

\bibitem{openai_o3_mini}
OpenAI.
\newblock Openai o3 mini, n.d.
\newblock Accessed: 2025-02-23.

\bibitem{team2023gemini}
Gemini Team, Rohan Anil, Sebastian Borgeaud, Jean-Baptiste Alayrac, Jiahui Yu, Radu Soricut, Johan Schalkwyk, Andrew~M Dai, Anja Hauth, Katie Millican, et~al.
\newblock Gemini: a family of highly capable multimodal models.
\newblock {\em arXiv preprint arXiv:2312.11805}, 2023.

\bibitem{wang2025baichuan}
Bingning Wang, Haizhou Zhao, Huozhi Zhou, Liang Song, Mingyu Xu, Wei Cheng, Xiangrong Zeng, Yupeng Zhang, Yuqi Huo, Zecheng Wang, et~al.
\newblock Baichuan-m1: Pushing the medical capability of large language models.
\newblock {\em arXiv preprint arXiv:2502.12671}, 2025.

\end{thebibliography}


\clearpage
\section{Data \& Code availability}

All our data, code, and generated responses from various models can be found in \href{https://github.com/MAGIC-AI4Med/MedRBench}{MedR-Bench}.


\section{Acknowledgments}
This work is supported by National Key R\&D Program of China (No. 2022ZD0160702).

\section{Author Contributions}
All listed authors clearly meet the ICMJE 4 criteria. P.Q. and C.W. contribute equally to this work. Y.W. and W.X. are the corresponding authors. Specifically, P.Q, C.W., S.L., W.Z., Z.C., H.G., C.P.,Y.Z., Y.W., and W.X. all make contributions to the conception or design of the work, and P.Q., C.W. further perform acquisition, analysis, or interpretation of data for the work. In writing,  P.Q. and C.W. draft the work. S.L., W.Z., Z.C., H.G., C.P.,Y.Z., Y.W., and W.X. review it critically for important intellectual content. All authors approve of the version to be published and agree to be accountable for all aspects of the work to ensure that questions related to the accuracy or integrity of any part of the work are appropriately investigated and resolved.


\clearpage
\section{Extended Figures}
\setcounter{table}{0}   
\setcounter{figure}{0}
\renewcommand{\figurename}{Extended Figure}
\begin{figure}[H]
    \centering
    \includegraphics[width=\linewidth]{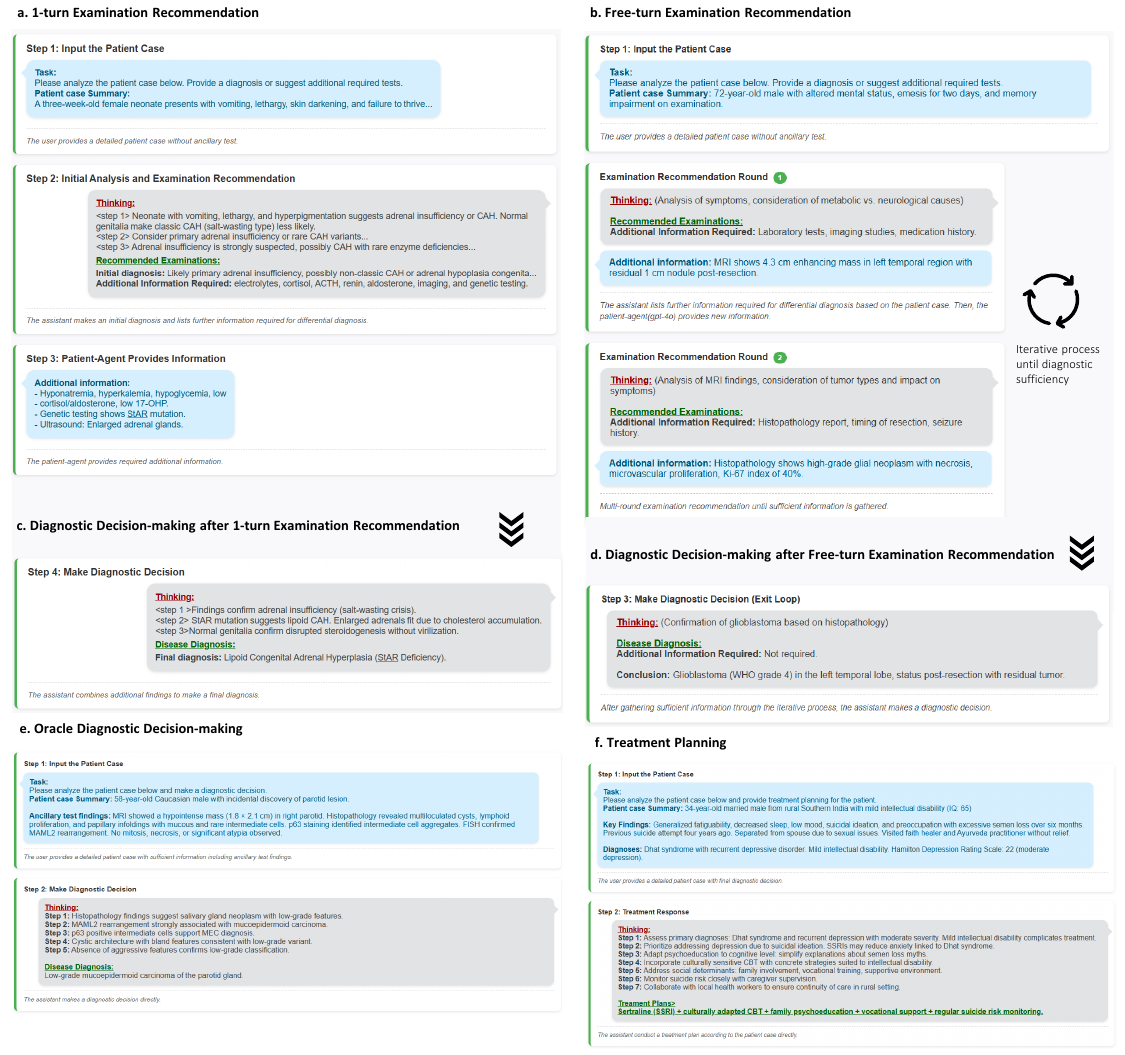}
    \caption{\textbf{Overview of our evaluation settings.} We consider three stages: examination recommendation, diagnosis decision-making, and treatment planning. \textbf{a, b} illustrate the 1-turn and free-turn interaction pipelines for examination recommendation. \textbf{c, d, e} depict the evaluation cases for diagnosis decision-making in 1-turn, free-turn, and oracle settings. Finally, \textbf{f} presents the treatment planning task in the oracle setting. }
    \label{fig:settings_show}
\end{figure}

\clearpage
\section{Extended Tables}
\renewcommand{\tablename}{Extended Table}

\begin{table}[H]
\tabcolsep=0.09cm
\renewcommand{\arraystretch}{1.2} 
\footnotesize
 \centering
\caption{\textbf{Results for {recommended examinations} on both ``all diseases'' and ``rare diseases'' in 1-turn and free-turn settings.}  The 0.95 confidence interval (CI) is reported in brackets.}
\label{tab:assessment_results}
\resizebox{\textwidth}{!}{
\begin{tabular}{l|c|cccc|cccc}
\toprule
\multicolumn{1}{c|}{\multirow{3}{*}{Method}}                        & \multicolumn{1}{c|}{\multirow{3}{*}{Model Size}}   & \multicolumn{4}{c|}{All Diseases}      & \multicolumn{4}{c}{Rare Diseases}    \\ \cline{3-10}

\multicolumn{1}{c|}{}                        & \multicolumn{1}{c|}{}    & \multicolumn{2}{c|}{1-turn}      & \multicolumn{2}{c|}{Free-turn}   & \multicolumn{2}{c|}{1-turn}      & \multicolumn{2}{c}{Free-turn}                      \\ \cline{3-10}

\multicolumn{1}{c|}{}                        & \multicolumn{1}{c|}{}                     &        Precision & \multicolumn{1}{c|}{Recall}  &{Precision} & \multicolumn{1}{c|}{Recall}  & Precision & \multicolumn{1}{c|}{Recall}  &{Precision} & \multicolumn{1}{c}{Recall}\\ \cline{1-10}
\rowcolor{mygray} \multicolumn{10}{c}{Close-source Reasoning LLMs} \\ \hline
OpenAI-o3-mini & --   & \makecell{33.75 \\(32.21, 35.29)}          & \makecell{38.47 \\(36.85, 40.10)}                          &      \makecell{33.57 \\(31.96, 35.19)}     &             \makecell{38.22 \\(36.59, 39.85)}  & \makecell{33.73 \\(31.60, 35.86)}           &\makecell{38.25 \\(36.03, 40.47)}                             &    \makecell{32.94 \\(30.77, 35.11)}                          &           \makecell{37.24 \\(35.04, 39.45)}      \\
 Gemini-2.0-FT   &     --          & \makecell{22.77 \\(21.42, 24.12)}          &\makecell{43.12 \\(41.41, 44.83)}                             &         \makecell{24.09 \\(22.54, 25.63)}     &             \makecell{39.88 \\(38.21, 41.55)}      & \makecell{21.93 \\(20.21, 23.64)}           &\makecell{42.96 \\(40.56, 45.36)}                             &             
 \makecell{22.73 \\(20.72, 24.75)}                          &           \makecell{39.66 \\(37.30, 42.01)}             \\\hline
\rowcolor{mygray} \multicolumn{10}{c}{Open-source Reasoning LLMs} \\ \hline
 DeepSeek-R1  & 671B  &\makecell{31.50 \\(29.97, 33.03)}  & \textbf{\makecell{43.61 \\(41.92, 45.30)}}     &       \makecell{32.23 \\(30.58, 33.89)}     &             \makecell{40.67 \\(38.92, 42.42)}               & \makecell{30.92 \\(28.81, 33.03)}  & \textbf{\makecell{43.73 \\(41.42, 46.04)}}    &   \makecell{31.64 \\(29.36, 33.93)}                          &           \makecell{40.18 \\(37.77, 42.59)}     \\
Qwen-QwQ   &  32B   &\makecell{24.43 \\(23.19, 25.67)}           &\makecell{39.90 \\(38.22, 41.57)}        & \makecell{25.39 \\(24.16, 26.63)}     &             \textbf{\makecell{40.83 \\(39.15, 42.51)}} & \makecell{24.08 \\(22.36, 25.80)}           &\makecell{39.44 \\(37.15, 41.74)}        &\makecell{25.47 \\(23.76, 27.18)}                          &           \textbf{\makecell{40.79 \\(38.51, 43.08)}} \\
Baichuan-M1    &    14B    & \textbf{\makecell{41.78 \\(39.91, 43.64)}}    &\makecell{37.88 \\(36.31, 39.45)}      &    \textbf{\makecell{41.99 \\(40.11, 43.87)}}    &             \makecell{36.99 \\(35.40, 38.59)}  &  \textbf{\makecell{41.58 \\(39.05, 44.12)}}   &\makecell{38.09 \\(36.01, 40.18)}      &     \textbf{\makecell{42.06 \\(39.57, 44.56)}}                         &           \makecell{37.73 \\(35.58, 39.89)}       \\ \bottomrule
\end{tabular}}
\end{table}

\begin{table}[H]
\renewcommand{\arraystretch}{1.2} 
\footnotesize
 \centering
\caption{\textbf{Results for {examination recommendation} reasoning on both ``all diseases'' and ``rare diseases'' in the 1-turn  settings.} For \textbf{DeepSeek-R1}, we assess its two types of reasoning, \emph{i.e.}, those presented in the formal answer part and the default thinking part, recorded as ``XX\scalebox{0.7}{ /XX}'' respectively. The 0.95 confidence interval (CI) is reported in brackets.}
\label{tab:assessment_reasoning}
\resizebox{.69\textwidth}{!}{
\begin{tabular}{l|c|cc|cc}
\toprule
\multicolumn{1}{c|}{\multirow{2}{*}{Method}}                        & \multicolumn{1}{c|}{\multirow{2}{*}{Model Size}}   & \multicolumn{2}{c|}{All Diseases}      & \multicolumn{2}{c}{Rare Diseases}    \\ \cline{3-6}

\multicolumn{1}{c|}{}                        & \multicolumn{1}{c|}{}                     &        Efficiency  & \multicolumn{1}{c|}{Factuality}   &{Efficiency} & \multicolumn{1}{c}{Factuality}\\ \cline{1-6}
\rowcolor{mygray} \multicolumn{6}{c}{Close-source Reasoning LLMs} \\ \hline
OpenAI-o3-mini & --      &     \makecell{95.17 \\(94.47, 95.87)}            &     \makecell{95.89 \\(95.18, 96.61)}                        &    \makecell{92.73 \\(91.16, 94.29)}&  \makecell{95.90 \\(94.93, 96.87)}      \\
 Gemini-2.0-FT   &     --                               &            \makecell{95.42 \\(94.79, 96.05)}         &      \textbf{\makecell{ 98.75 \\(98.38, 99.13)} }                       &            
\makecell{84.22 \\(82.62, 85.83)}&            \textbf{\makecell{98.98 \\(98.52, 99.43)}}  \\\hline
\rowcolor{mygray} \multicolumn{6}{c}{Open-source Reasoning LLMs} \\ \hline
 DeepSeek-R1  & 671B     &       \textbf{\makecell{98.59 \scalebox{0.7}{/89.75}\\(98.17, 99.00)}}        &    \makecell{96.79 \scalebox{0.7}{/95.38}\\(96.17, 97.41)}               &   \textbf{\makecell{95.96 \scalebox{0.7}{/88.91}\\(94.83, 97.08)}}  &  \makecell{97.02 \scalebox{0.7}{/95.10}\\(96.17, 97.87)}     \\
Qwen-QwQ   &  32B       & \makecell{86.53 \\(85.34, 87.71)}&\makecell{94.87 \\(94.09, 95.64)}        &\makecell{76.34 \\(74.02, 78.67)}     &  \makecell{94.53 \\(93.45, 95.62)}\\
Baichuan-M1    &    14B       &    \makecell{90.66 \\(89.66, 91.66)}    &    \makecell{96.87 \\(96.35, 97.39)}      &     \makecell{84.37 \\(82.26, 86.48)}       &      \makecell{97.06 \\(96.36, 97.76)}    \\ \bottomrule
\end{tabular}}
\end{table}

\begin{table}[H]
\tabcolsep=0.1cm
\renewcommand{\arraystretch}{1.2} 
\footnotesize
\centering
\caption{\textbf{Results for disease diagnosis in the 1-turn, free-turn, and oracle evaluation settings on both ``all diseases'' and ``rare diseases''.} The 0.95 confidence interval (CI) is reported in brackets.}
\label{tab:diagnostic_results}
\resizebox{0.8\textwidth}{!}{
\begin{tabular}{l|c|ccc|ccc}
\toprule
\multicolumn{1}{c|}{\multirow{3}{*}{Method}} &
\multicolumn{1}{c|}{\multirow{3}{*}{Model Size}} &
\multicolumn{3}{c|}{All Diseases} &
\multicolumn{3}{c}{Rare Diseases} \\ \cline{3-8}

\multicolumn{1}{c|}{} & \multicolumn{1}{c|}{} &
\multicolumn{1}{c|}{1-turn} &
\multicolumn{1}{c|}{Free-turn} &
\multicolumn{1}{c|}{Oracle} &
\multicolumn{1}{c|}{1-turn} &
\multicolumn{1}{c|}{Free-turn} &
\multicolumn{1}{c}{Oracle} \\  \cline{3-8}

\multicolumn{1}{c|}{} & \multicolumn{1}{c|}{} &
\multicolumn{1}{c|}{Accuracy} &
\multicolumn{1}{c|}{Accuracy} &
\multicolumn{1}{c|}{Accuracy} &
\multicolumn{1}{c|}{Accuracy} &
\multicolumn{1}{c|}{Accuracy} &
\multicolumn{1}{c}{Accuracy} \\ \hline

\rowcolor{mygray} \multicolumn{8}{c}{Close-source Reasoning LLMs} \\ \hline
OpenAI-o3-mini &
-- &
\makecell{64.99 \\(61.97, 68.02)} &
\makecell{67.19 \\(64.21, 70.17)} &
\makecell{83.91 \\(81.58, 86.24)} &
\makecell{63.75 \\(59.48, 68.01)} &
\makecell{65.99 \\(61.78, 70.19)} &
\makecell{85.54 \\(82.43, 88.65)}  \\

Gemini-2.0-FT &
-- &
\makecell{68.55 \\(65.60, 71.49)} &
\makecell{68.76 \\(65.81, 71.70)} &
\makecell{86.83 \\(84.69, 88.98)} &
\makecell{66.40 \\(62.20, 70.59)} &
\makecell{68.43 \\(64.31, 72.56)} &
\makecell{88.60 \\(85.78, 91.41)} \\ \hline

\rowcolor{mygray} \multicolumn{8}{c}{Open-source Reasoning LLMs} \\ \hline
DeepSeek-R1 &
671B &
\textbf{\makecell{71.79 \\(68.93, 74.64)}} &
\textbf{\makecell{76.18 \\(73.47, 78.88)}} &
\textbf{\makecell{89.76 \\(87.84, 91.68)}} &
\textbf{\makecell{70.67 \\(66.63, 74.71)}} &
\textbf{\makecell{77.60 \\(73.90, 81.30)}} &
\textbf{\makecell{91.04 \\(88.51, 93.57)}} \\

Qwen-QwQ &
32B &
\makecell{63.74 \\(60.69, 66.79)} &
\makecell{74.71 \\(71.95, 77.47)} &
\makecell{85.06 \\(82.80, 87.32)} &
\makecell{64.15 \\(59.90, 68.41)} &
\makecell{73.93 \\(70.03, 77.83)} &
\makecell{86.15 \\(83.09, 89.21)} \\

Baichuan-M1 &
14B &
\makecell{66.77 \\(63.78, 69.76)} &
\makecell{61.23 \\(58.14, 64.33)} &
\makecell{84.43 \\(82.13, 86.73)} &
\makecell{67.21 \\(63.04, 71.38)} &
\makecell{60.69 \\(56.36, 65.03)} &
\makecell{88.39 \\(85.55, 91.23)} \\ \bottomrule
\end{tabular}}
\end{table}

\clearpage
\begin{table}[tbp!]
\tabcolsep=0.09cm
\renewcommand{\arraystretch}{1.2} 
\footnotesize
\centering
\caption{\textbf{Results for diagnostic decision-making reasoning in the 1-turn and oracle evaluation settings on ``all diseases''.} For \textbf{DeepSeek-R1}, we assess its two types of reasoning, \emph{i.e.}, those presented in the formal answer part and the default thinking part, recorded as ``XX\scalebox{0.7}{ /XX}'' respectively. The 0.95 confidence interval (CI) is reported in brackets.}
\label{tab:diagnostic_reasoning_all_diseases}
\resizebox{0.8\textwidth}{!}{
\begin{tabular}{l|c|ccc|ccc}
\toprule
\multicolumn{1}{c|}{\multirow{2}{*}{Method}} &
\multicolumn{1}{c|}{\multirow{2}{*}{Model Size}} &
\multicolumn{3}{c|}{1-turn reasoning} &
\multicolumn{3}{c}{Oracle reasoning} \\ \cline{3-8}
\multicolumn{1}{c|}{} & \multicolumn{1}{c|}{} &
\multicolumn{1}{c}{Efficiency} &
\multicolumn{1}{c}{Factuality} &
\multicolumn{1}{c|}{Completeness} &
\multicolumn{1}{c}{Efficiency} &
\multicolumn{1}{c}{Factuality} &
\multicolumn{1}{c}{Completeness} \\ \hline

\rowcolor{mygray} \multicolumn{8}{c}{Close-source Reasoning LLMs} \\ \hline
OpenAI-o3-mini &
-- &
\makecell{91.59 \\(90.37, 92.81)} &
\makecell{83.15 \\(81.29, 85.01)} &
\makecell{50.87 \\(48.60, 53.13)} &
\makecell{94.33 \\(93.61, 95.05)} &
\makecell{94.94 \\(94.23, 95.64)} &
\makecell{75.42 \\(74.14, 76.70)} \\

Gemini-2.0-FT &
-- &
\makecell{83.77 \\(82.56, 84.97)} &
\makecell{87.17 \\(85.52, 88.83)} &
\makecell{54.45 \\(52.20, 56.69)} &
\makecell{95.89 \\(95.35, 96.43)} &
\textbf{\makecell{98.23 \\(97.84, 98.62)}} &
\textbf{\makecell{83.28 \\(82.17, 84.40)}} \\ \hline

\rowcolor{mygray} \multicolumn{8}{c}{Open-source Reasoning LLMs} \\ \hline
DeepSeek-R1 &
671B &
\textbf{\makecell{95.86\scalebox{0.7}{/88.24} \\(95.02, 96.71)}} &
\makecell{87.15\scalebox{0.7}{/85.90} \\(85.54, 88.76)} &
\makecell{54.88\scalebox{0.7}{/67.59} \\(52.77, 56.99)} &
\textbf{\makecell{97.17\scalebox{0.7}{/89.73} \\(96.65, 97.70)}} &
\makecell{95.03\scalebox{0.7}{/96.31} \\(94.34, 95.71)} &
\makecell{78.27\scalebox{0.7}{/90.79} \\(77.17, 79.37)} \\

Qwen-QwQ &
32B &
\makecell{76.97 \\(75.30, 78.64)} &
\makecell{88.14 \\(86.48, 89.80)} &
\textbf{\makecell{66.94 \\(63.83, 70.04)}} &
\makecell{71.20 \\(69.46, 72.94)} &
\makecell{84.02 \\(82.23, 85.80)} &
\makecell{79.97 \\(78.11, 81.83)} \\

Baichuan-M1 &
14B &
\makecell{82.91 \\(81.34, 84.48)} &
\textbf{\makecell{88.62 \\(87.23, 90.01)}} &
\makecell{53.43 \\(51.17, 55.68)} &
\makecell{92.80 \\(92.09, 93.51)} &
\makecell{96.84 \\(96.35, 97.32)} &
\makecell{75.11 \\(73.87, 76.36)} \\ \bottomrule
\end{tabular}}
\end{table}

\begin{table}[!htb]
\tabcolsep=0.09cm
\renewcommand{\arraystretch}{1.2} 
\footnotesize
\centering
\caption{\textbf{Results for diagnostic decision-making reasoning in the 1-turn and oracle evaluation settings on ``rare diseases''.} For \textbf{DeepSeek-R1}, we assess its two types of reasoning, \emph{i.e.}, those presented in the formal answer part and the default thinking part, recorded as ``XX\scalebox{0.7}{ /XX}'' respectively. The 0.95 confidence interval (CI) is reported in brackets.}
\label{tab:diagnostic_reasoning_rare_diseases}
\resizebox{0.8\textwidth}{!}{
\begin{tabular}{l|c|ccc|ccc}
\toprule
\multicolumn{1}{c|}{\multirow{2}{*}{Method}} &
\multicolumn{1}{c|}{\multirow{2}{*}{Model Size}} &
\multicolumn{3}{c|}{1-turn reasoning} &
\multicolumn{3}{c}{Oracle reasoning} \\ \cline{3-8}
\multicolumn{1}{c|}{} & \multicolumn{1}{c|}{} &
\multicolumn{1}{c}{Efficiency} &
\multicolumn{1}{c}{Factuality} &
\multicolumn{1}{c|}{Completeness} &
\multicolumn{1}{c}{Efficiency} &
\multicolumn{1}{c}{Factuality} &
\multicolumn{1}{c}{Completeness} \\ \hline

\rowcolor{mygray} \multicolumn{8}{c}{Close-source Reasoning LLMs} \\ \hline
OpenAI-o3-mini &
-- &
\makecell{92.73 \\(91.16, 94.29)} &
\makecell{81.28 \\(78.58, 83.99)} &
\makecell{49.83 \\(46.59, 53.07)} &
\makecell{94.80 \\(93.83, 95.77)} &
\makecell{95.02 \\(94.14, 95.89)} &
\makecell{76.44 \\(74.74, 78.14)} \\

Gemini-2.0-FT &
-- &
\makecell{84.22 \\(82.62, 85.83)} &
\makecell{85.93 \\(83.53, 88.33)} &
\makecell{54.06 \\(50.85, 57.28)} &
\makecell{96.45 \\(95.74, 97.16)} &
\textbf{\makecell{98.39 \\(97.89, 98.89)}} &
\textbf{\makecell{84.30 \\(82.75, 85.85)}} \\ \hline

\rowcolor{mygray} \multicolumn{8}{c}{Open-source Reasoning LLMs} \\ \hline
DeepSeek-R1 &
671B &
\textbf{\makecell{95.96\scalebox{0.7}{/88.91} \\(94.83, 97.08)}} &
\makecell{84.76\scalebox{0.7}{/84.70} \\(82.30, 87.22)} &
\makecell{54.10\scalebox{0.7}{/67.54} \\(51.05, 57.15)} &
\textbf{\makecell{97.61\scalebox{0.7}{/90.33} \\(96.95, 98.27)}} &
\makecell{94.75\scalebox{0.7}{/96.83} \\(93.77, 95.74)} &
\makecell{79.00\scalebox{0.7}{/91.14} \\(77.52, 80.48)} \\

Qwen-QwQ &
32B &
\makecell{76.34 \\(74.02, 78.67)} &
\textbf{\makecell{89.05 \\(86.79, 91.31)}} &
\textbf{\makecell{66.53 \\(62.14, 70.92)}} &
\makecell{72.25 \\(69.83, 74.68)} &
\makecell{84.30 \\(81.83, 86.77)} &
\makecell{80.70 \\(78.16, 83.24)} \\

Baichuan-M1 &
14B &
\makecell{84.37 \\(82.26, 86.48)} &
\makecell{88.90 \\(87.06, 90.73)} &
\makecell{53.11 \\(49.95, 56.27)} &
\makecell{93.94 \\(93.07, 94.81)} &
\makecell{96.91 \\(96.26, 97.56)} &
\makecell{76.05 \\(74.31, 77.78)} \\ \bottomrule
\end{tabular}}
\end{table}

\begin{table}[!htb]
\tabcolsep=0.1cm
\renewcommand{\arraystretch}{1.2} 
\footnotesize
 \centering
\caption{\textbf{Results for {treatment planning} on both ``all diseases'' and ``rare diseases''.} We calculate metrics for both the final generation and reasoning processes. For \textbf{DeepSeek-R1}, we have assessed its two types of reasoning, \emph{i.e.}, those presented in the formal answer part and the default thinking part, recorded as ``XX\scalebox{0.7}{ /XX}'' respectively. The 0.95 confidence interval (CI) is reported in brackets.}
\label{tab:treatment}
\resizebox{\textwidth}{!}{
\begin{tabular}{l|c|cccc|cccc}

 \toprule
\multicolumn{1}{c|}{\multirow{3}{*}{Method}} & \multicolumn{1}{c|}{\multirow{3}{*}{Model Size}} & \multicolumn{4}{c|}{All Diseases}                               & \multicolumn{4}{c}{Rare Diseases}                                \\ \cline{3-10} 
\multicolumn{1}{c|}{}                        & \multicolumn{1}{c|}{}                            & \multicolumn{1}{c|}{Treatment Plans} & \multicolumn{3}{c|}{Reasoning Processes}         & \multicolumn{1}{c|}{Treatment Plans} & \multicolumn{3}{c}{Reasoning Processes}          \\ \cline{3-10} 
\multicolumn{1}{c|}{}                        & \multicolumn{1}{c|}{}                            & \multicolumn{1}{c|}{Accuracy}     & Efficiency & Factuality & Completeness & \multicolumn{1}{c|}{Accuracy}     & Efficiency & Factuality & Completeness \\ \hline
\rowcolor{mygray} \multicolumn{10}{c}{Close-source Reasoning LLMs} \\ \hline
OpenAI-o3-mini & -- & \makecell{27.03 \\ (23.09, 30.97)} & \makecell{94.67 \\(93.90, 95.44)} & \makecell{96.77 \\(96.06, 97.48)}& \makecell{70.63 \\(68.55, 72.71)}& \makecell{23.17 \\ (16.65, 29.70)} &\makecell{95.06 \\(93.79, 96.33)}& \textbf{\makecell{96.81 \\(95.51, 98.11)}}&\makecell{69.86 \\(66.33, 73.38)}\\
Gemini-2.0-FT  &    --   & \makecell{25.66 \\ (21.80, 29.52)} &  \makecell{93.66 \\(92.82, 94.49)}& \textbf{\makecell{96.96 \\(96.34, 97.59)}} & \makecell{75.89 \\(73.81, 77.98) }& \makecell{23.78 \\ (17.20, 30.37)} & \makecell{94.41 \\(93.07, 95.74)}&\makecell{96.68 \\(95.57, 97.78)}&\makecell{77.10 \\(73.68, 80.51)}\\ \hline
\rowcolor{mygray} \multicolumn{10}{c}{Open-source Reasoning LLMs} \\ \hline
 DeepSeek-R1                                         &  671B  & \makecell{30.51 \\ (26.43, 34.58)} & \textbf{\makecell{95.25 \scalebox{0.7}{/88.93}\\(94.43, 96.08)}}& \makecell{94.59 \scalebox{0.7}{/95.93}\\(93.63, 95.56)}& \makecell{68.08 \scalebox{0.7}{/80.48}\\(65.97, 70.20)}& \makecell{27.27 \\ (20.41, 34.14)} & \textbf{\makecell{ 95.37 \scalebox{0.7}{/89.19}\\(93.92, 96.82)}} &\makecell{95.22 \scalebox{0.7}{/95.01}\\(93.62, 96.47)}&\makecell{68.28 \scalebox{0.7}{/81.05}\\(64.57, 71.99)}\\
Qwen-QwQ     &    32B   &    \makecell{20.89 \\ (17.12, 24.66)} & \makecell{84.76 \\(83.42, 86.10)}& \makecell{94.40 \\(93.44, 95.36)}& \textbf{\makecell{77.66 \\(75.36, 79.96)}} & \makecell{19.61 \\ (13.25, 25.97)} &\makecell{83.31 \\(81.08, 85.54)}&\makecell{94.05 \\(92.00, 96.10)}& \textbf{\makecell{78.74 \\(74.85, 82.63)}}\\
Baichuan-M1      &     14B     &  \textbf{\makecell{30.65 \\ (26.57, 34.72)}} &    \makecell{88.47 \\(87.37, 89.57)}& \makecell{96.56 \\(95.80, 97.31)}& \makecell{ 70.18 \\(68.04, 72.32)}& \textbf{\makecell{30.30 \\ (23.22, 37.39)}} & \makecell{ 87.87 \\(85.96, 89.77)}& \makecell{95.97 \\(94.62, 97.31)}& \makecell{69.56 \\(65.91, 73.21)}\\ \bottomrule
\end{tabular}}
\end{table}

\clearpage



\section{Supplementary}
\setcounter{table}{0}   
\setcounter{figure}{0}
\renewcommand{\tablename}{Supplementary Table}
\renewcommand{\figurename}{Supplementary Figure}


\subsection{Qualitative Case Study}
\label{case_study}

In this part, we will analyze the performance of various models qualitatively.

For clarity, we first elucidate the meanings of the terms in the leftmost column of the case study table, as follows.
\begin{itemize}
    \item \textbf{Case ID:} indicates the unique identifier of the PMC case report. 
    
    \item \textbf{Category:} introduces the classifications of this case, including body system, disorders and conditions, and whether it is rare disease-related. 
    
    \item \textbf{Case Summary:} provides the basic patient information excluding ancillary tests.  
    
    \item \textbf{Differential Diagnosis:} describes the ground-truth diagnostic process extracted directly from the case report. 
    
    \item \textbf{Final Diagnosis:} states the basic factual diagnosis of the patient’s case and related statements summarized from the original case report. 
    
    \item \textbf{Diagnosis Result:} is the name of the diagnosed disease. 
    In Case 1 for examination recommendation.
    
    \item \textbf{Ancillary Tests Split:} presents the ground-truth ancillary test information structured by GPT-4o.
    
    \item \textbf{Treatment Planning Analysis:} presents the ground-truth rationale for the preferred treatment plan~(case 4).
    
    \item \textbf{Treatment Plan Results:} describes the preferred treatment plan itself. 
\end{itemize}










All of the above are the basic patient information, as well as the real diagnosis results or treatment plans, extracted and organized from the case report through the model.

The following rows show responses from the LLMs.
For Case 2, Case 3, and Case 4, in order to facilitate the comparison between the reasoning process of the models under evaluation for diagnostic decision-making or treatment planning and the ground truth reasoning process extracted from the case report, we present the three side by side. However, for Case 1, which involves the examination recommendation task, the original case report typically does not specify the reasoning process; therefore, we do not display the ground truth column here.  In Case 1 for examination recommendation, 

\begin{itemize}
    \item \textbf{Request Reasoning:} represents the reasoning process during the examination recommendation procedure. 

    \item \textbf{Request Output:} indicates the additional ancillary test information requested by the LLM. 

    \item \textbf{Ancillary Tests Provided:} shows the supplementary information furnished in response. With this additional information, the model proceeds to make a diagnostic decision. 
\end{itemize}




For the first 3 cases that include the diagnostic process, 
\begin{itemize}
    \item \textbf{Diagnostic Reasoning:} presents the reasoning process during the diagnostic procedure.
    
    \item \textbf{Request Split:} provides the structured version of \textbf{Request output}, which is used to calculate precision and recall. 
    
    \item \textbf{Diagnosis:} presents the final diagnosis determined by the LLM. 
    
\end{itemize}




In Case4, we compare the differences in reasoning procedures between Deepseek-R1's chain of thought and its thinking process. 

\begin{itemize}
    \item \textbf{Treatment Reasoning:} presents the reasoning process during the treatment planning procedure.The ground truth of reasoning steps is a reformatted version of the \textbf{treatment planning analysis} that presents the rationale in a step-by-step format. 

    \item \textbf{Predicted Treatment Plan:} indicates the treatment plan selected by the LLMs. 
\end{itemize}



For all cases, the definitions of metrics such as Accuracy, Precision, and Recall are consistent with those provided in the methods section.

\subsubsection{Examination Recommendation}
Here, we present a case study that vividly illustrates the practical implementation of the 1-turn examination recommendation process. This case, as shown in Supplementary Figure~\ref{fig:case1}, shows how LLMs perform in terms of symptom identification and actively querying information. For the sake of presenting results concisely, we only show the response of one open-source model (DeepSeek-R1) and one closed-source model (OpenAI-o3-mini).

\noindent \textbf{Initial Analysis.} 

In the initial phase of examination recommendation, encompassing the preliminary analysis and subsequent information inquiry, the models exhibited robust performance. They effectively identified pivotal symptoms, including vomiting, lethargy, skin darkening, and failure to thrive, utilizing these indicators to guide further information requests. The models demonstrated an understanding that specific symptom clusters were indicative of adrenal pathology, subsequently enumerating potential diagnoses such as congenital adrenal hyperplasia (CAH), Addison’s disease, and metabolic disorders. Building upon these tentative diagnoses, the models conducted a comprehensive assessment of the presenting symptoms to facilitate preliminary exclusion. For instance, DeepSeek-R1 ruled out neuroblastoma or other neoplastic conditions based on the absence of abdominal masses in the patient. In cases where definitive exclusion was not possible with the available information, the models considered requisite diagnostic tests for further examination. Regarding the presumptive diagnosis of CAH, the models contemplated relevant information that could refine diagnostic accuracy. Notably, they recognized the significance of normal genitalia as a critical piece of information, challenging the typical clinical presentation of CAH and prompting a more nuanced diagnostic approach.

\noindent \textbf{Request for Additional Examinations.} 

Based on the initial analysis, the models listed a comprehensive range of laboratory tests, including serum electrolytes, cortisol, adrenocorticotropic hormone (ACTH), 17-hydroxyprogesterone, renin, aldosterone levels, and blood glucose, imaging examinations (abdominal ultrasound), as well as genetic testing and urine steroid profile analysis. This underscores the model’s proficiency in identifying the requisite diagnostic tests to differentiate among the various proposed diagnoses. Notably, during the evaluation phase, when contemplating Congenital Lipoid Adrenal Hyperplasia (CLAH), the model specifically solicited urine steroid profiles or genetic testing to ascertain the presence of adrenal enzyme deficiencies, thereby demonstrating a nuanced understanding of the specific disease and its features.

\begin{figure}
    \centering
    \includegraphics[width=0.95\linewidth]{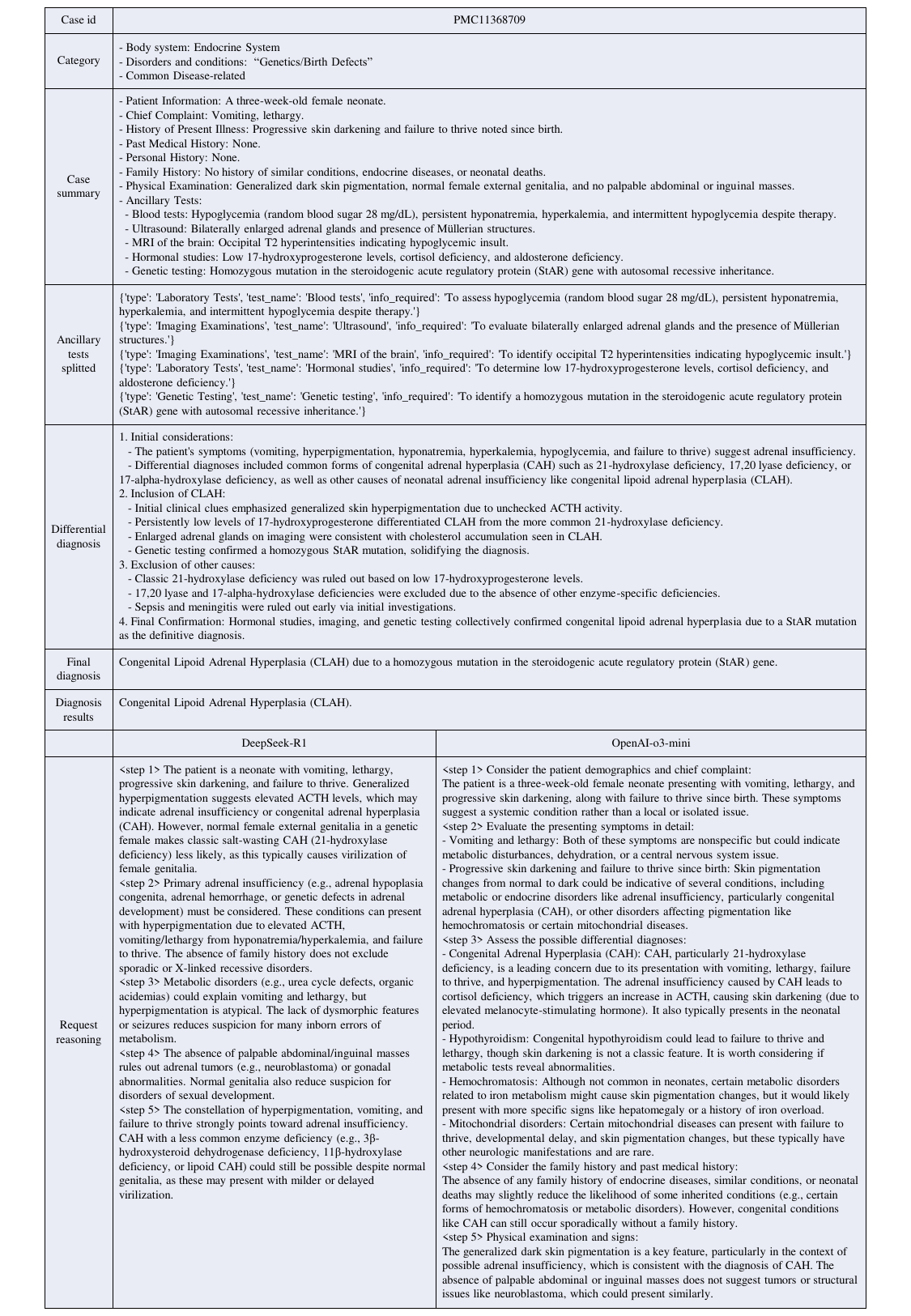}
\end{figure}
\begin{figure}
    \centering
    \includegraphics[width=0.95\linewidth]{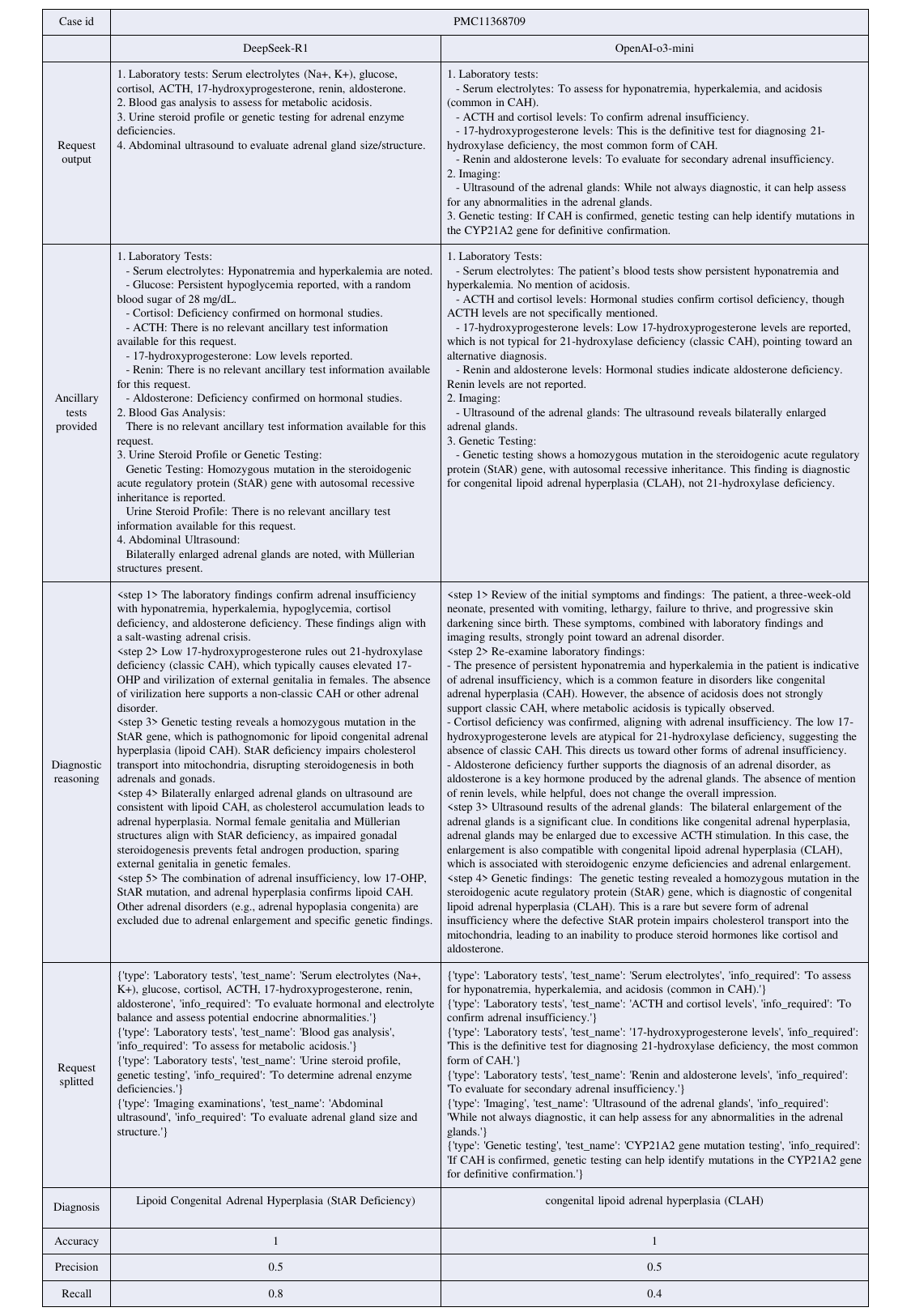}
    \caption{Case 1. A case of 1-turn examination recommendation and diagnostic decision-making. The meaning of the row headers is explained at the beginning of Supplementary~\ref{case_study}}
    \label{fig:case1}
\end{figure}

\subsubsection{Diagnostic Decision-making}
\begin{figure}
    \centering
    \includegraphics[width=0.95\linewidth]{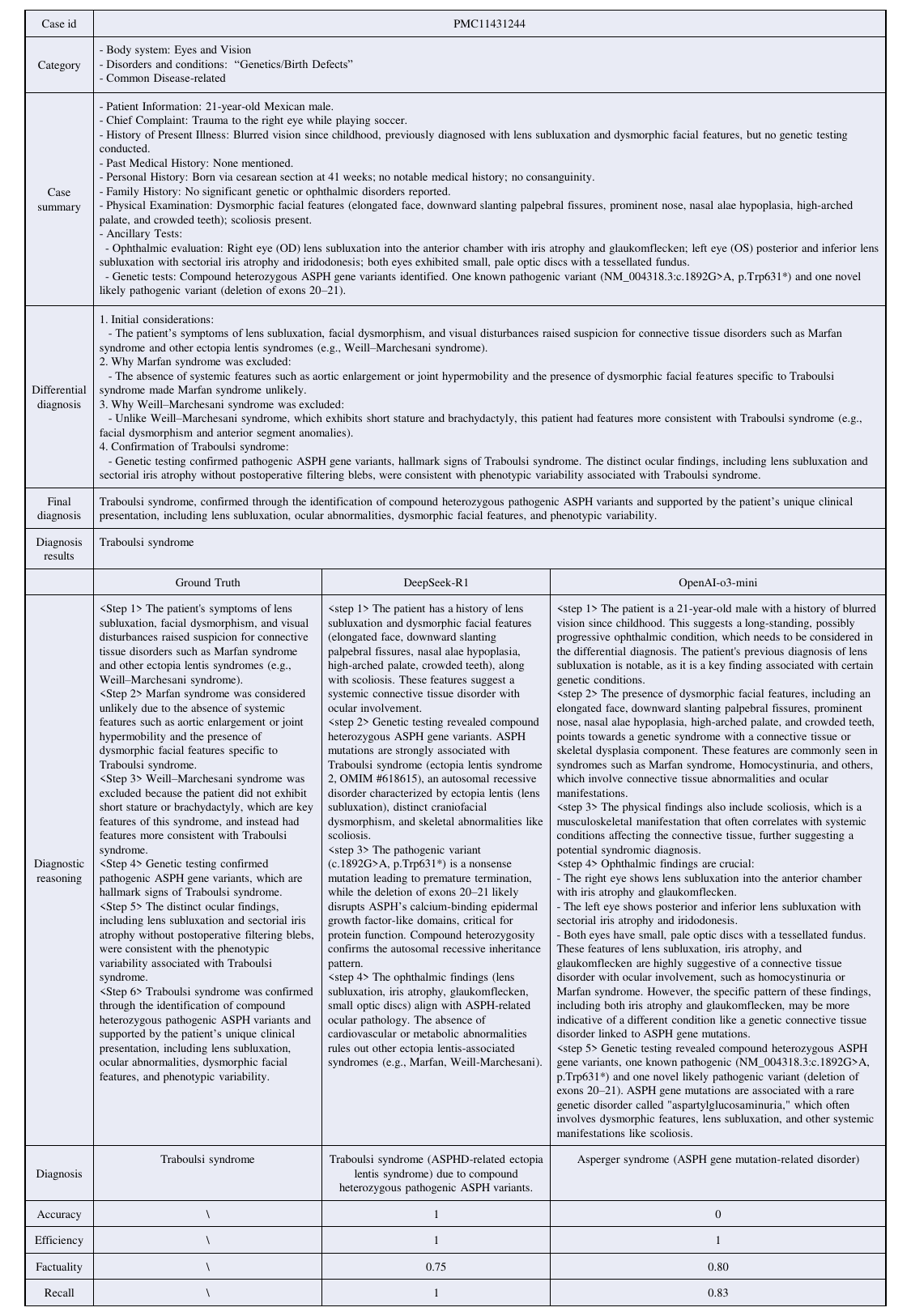}
    \caption{Case 2. A case of oracle diagnosis on common disease. The meaning of the row headers is explained at the beginning of Supplementary~\ref{case_study}}
    \label{fig:case2}
\end{figure}

\begin{figure}
    \centering
    \includegraphics[width=0.95\linewidth]{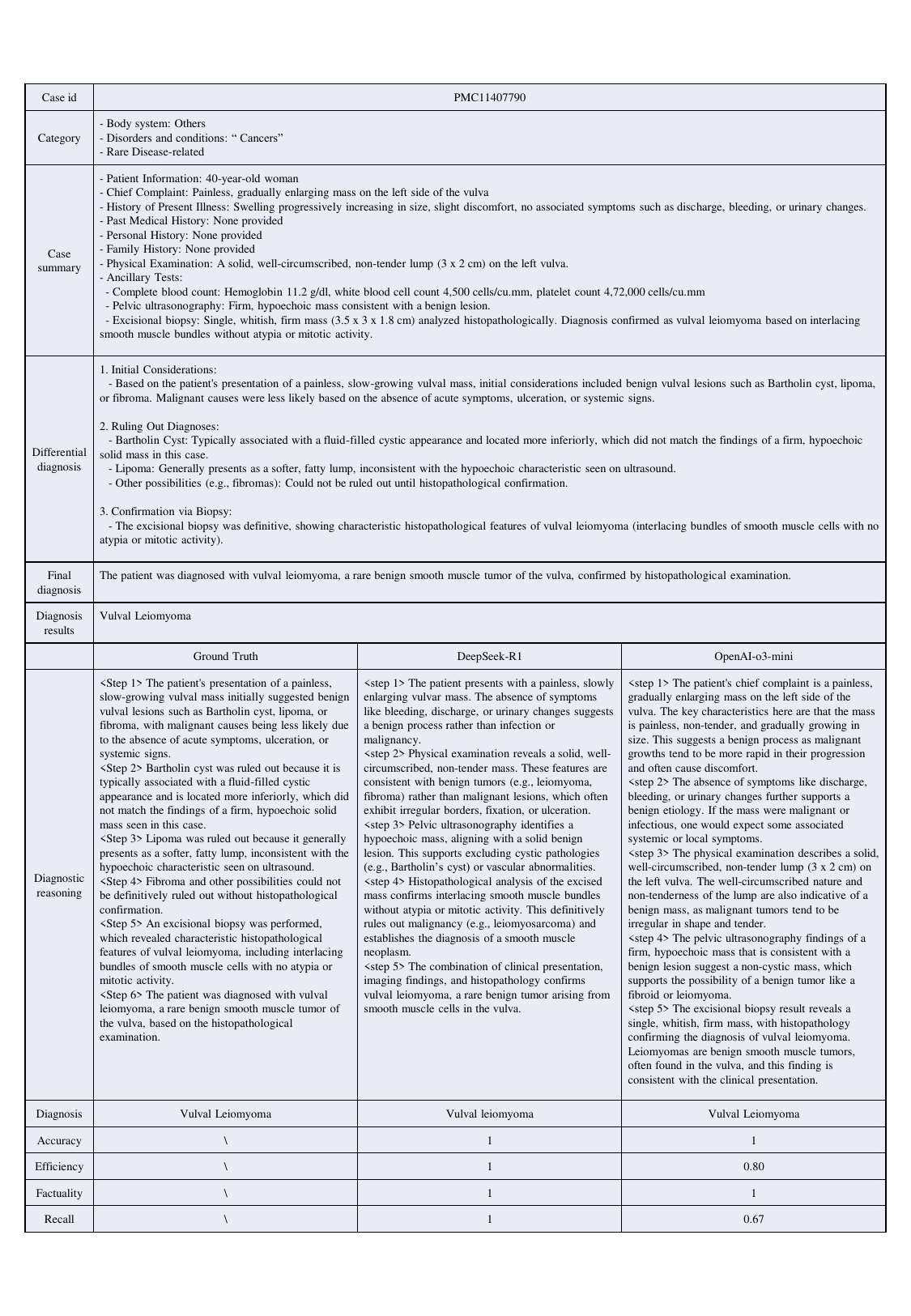}
    \caption{Case 3. A case of oracle diagnosis on rare diseases. The meaning of the row headers is explained at the beginning of Supplementary~\ref{case_study}}
    \label{fig:case3}
\end{figure}

In this case study, we explore the diagnostic reasoning processes employed by DeepSeek-R1 and OpenAI-o3-mini across two different settings: (1) Diagnostic decision-making after 1-turn examination recommendation (Supplementary Figure~\ref{fig:case1}). (2) Oracle diagnostic decision-making on a common disease case (Supplementary Figure~\ref{fig:case2}) and rare disease case (Supplementary Figure~\ref{fig:case3}).

\noindent \textbf{Diagnostic decision-making after 1-turn examination recommendation.}

When additional information was presented, the models carefully analyze each item. They noticed how the laboratory test results, such as hyponatremia, hyperkalemia, hypoglycemia, cortisol and aldosterone deficiencies, and low 17-hydroxyprogesterone, fit in with the previously considered diagnoses. It also took into account the genetic test result (homozygous mutation in the steroidogenic acute regulatory protein (StAR) gene) and the ultrasound finding (bilaterally enlarged adrenal glands) to further refine their understanding and inference. This shows that they could effectively integrate new information into the existing framework and adjust their thinking accordingly.

Initially, based on the initial symptoms and available information, the model further excluded or retained the possibilities proposed in the initial analysis. For example, during the diagnostic reasoning process, OpenAI-o3-mini further excluded classic CAH according to the absence of acidosis in the patient, as metabolic acidosis is usually observed in classic CAH. DeepSeek-R1 also excluded 21-hydroxylase deficiency (classic CAH) based on the low 17-hydroxyprogesterone levels and the absence of virilization of the genitalia. Then, they used the genetic test result and the ultrasound finding to confirm the diagnosis of lipoid congenital adrenal hyperplasia (CLAH). This indicates that the models could use information in a logical and evidence-based manner to reach the correct final diagnosis.

\noindent \textbf{Oracle diagnostic decision-making.}

When provided with complete information, LLMs tend to perform accurate diagnosis. In this context, our analysis focuses on evaluating the efficiency of reasoning, the factuality of each step, the completeness (recall of ground truth reasoning steps), and the final diagnostic accuracy.

\textbf{Efficiency:} DeepSeek-R1's approach is characterized by its conciseness and directness. In Case 2, it swiftly progresses from symptoms to targeting Traboulsi syndrome through genetic testing and clinical manifestations. Similarly, in Case 3, it efficiently eliminates malignant and infectious causes early on, focusing on mass characteristics and ultrasound findings for diagnosis. In contrast, OpenAI-o3-mini provided a more detailed step-by-step analysis. In Case 2, it involved more physical examinations and eye findings. In case 3, it emphasized the benign nature of the mass through the absence of symptoms typically associated with malignancy or infection. While this information contributes to a comprehensive understanding of the condition, it also adds complexity to reasoning. 

\textbf{Factuality:}
While generally aligned with medical knowledge, there are instances where the models deviate. In Case 2, DeepSeek-R1's reasoning step 4  is not strictly factual. The absence of cardiovascular or metabolic abnormalities is not a sufficient condition to exclude Marfan or Weill-Marchesani syndromes, which are characterized by specific symptoms like aortic enlargement or joint hypermobility, and short stature or brachydactyly, respectively. Meanwhile, OpenAI-o3-mini's reasoning step 5 assertion about ASPH gene mutations lacks medical support and may have significantly contributed to its incorrect final diagnosis. In Case 3, however, both models adhere closely to medical principles, with DeepSeek-R1 correctly associating symptom absence with a benign process and OpenAI-o3-mini accurately emphasizing the importance of mass characteristics in suggesting a benign etiology.

\textbf{Completeness:}
DeepSeek-R1 demonstrates a strong ability to recall and follow GT reasoning steps in both cases, particularly in identifying the key clues such as the benign nature of lesions and the importance of gene testing or histopathology in diagnosis. OpenAI-o3-mini, however, shows a limited capacity to explicitly exclude differential diagnoses. In Case 1, it failed to clearly rule out Weill-Marchesani syndrome, and in Case 2, it did not definitively exclude Bartholin cyst, despite its detailed step-by-step analysis.

\textbf{Diagnostic Accuracy:}
In terms of final diagnosis, both models correctly identify the ASPH mutation in Case 2. However, OpenAI-o3-mini misdiagnoses the condition as "Asperger Syndrome (ASPH gene mutation-related disorder)", which is a misinterpretation, as Asperger Syndrome is a pervasive developmental disorder not proven to be related to the ASPH gene. This indicates a gap in OpenAI-o3-mini's medical knowledge and reasoning rigor. In Case 3, both DeepSeek-R1 and OpenAI-o3-mini successfully diagnosed vulvar leiomyoma. This consistency with the GT suggests that both models are capable of accurate clinical reasoning when presented with clear symptoms and diagnostic findings.

\begin{figure}
    \centering
    \includegraphics[width=0.9\linewidth]{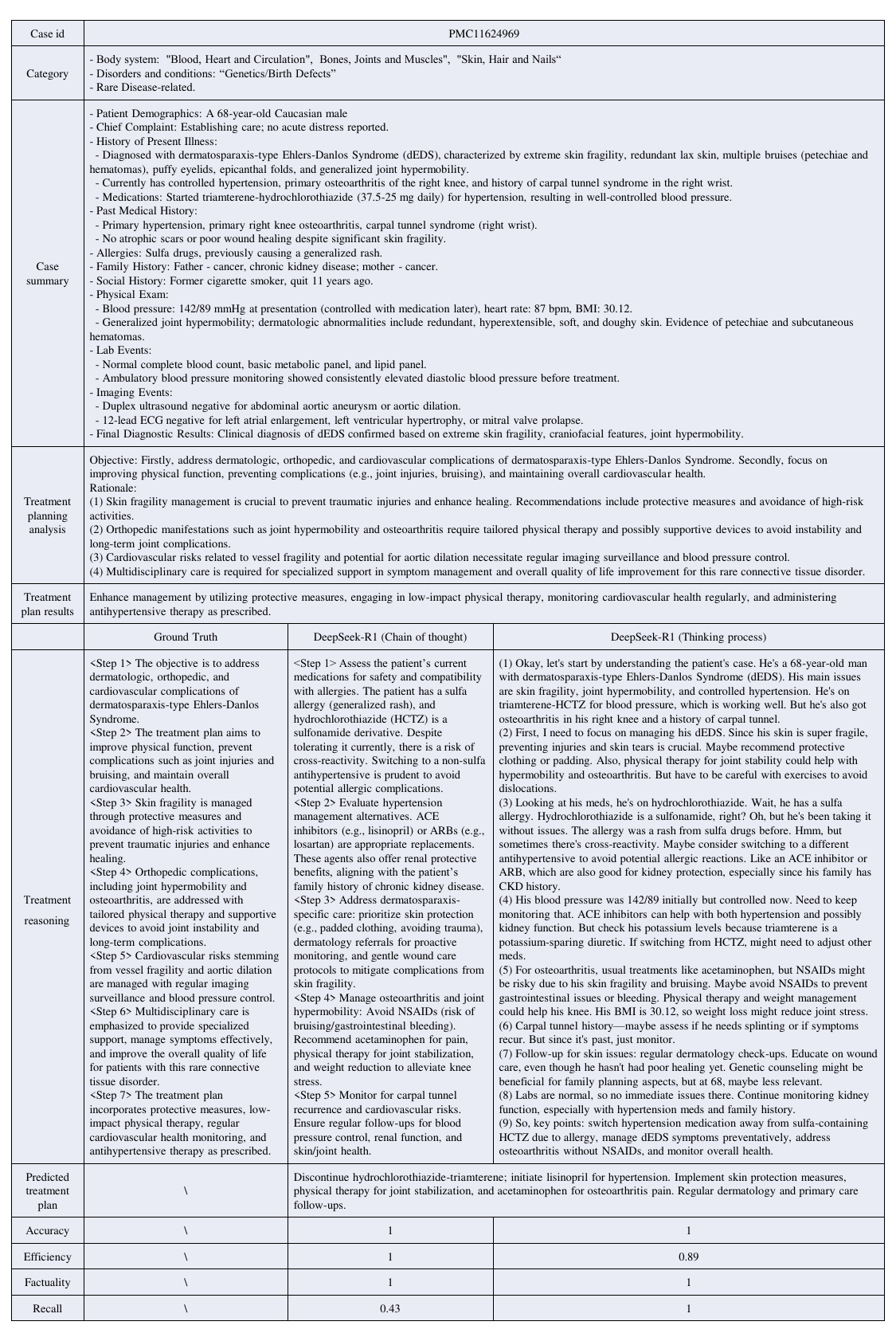}
    \caption{Case 4. A case of treatment planning for a rare disease. The meaning of the row headers is explained at the beginning of Supplementary~\ref{case_study}.
    }
    \label{fig:treament_planing_case1}
\end{figure}

\subsubsection{Treatment Planning}
Here, we present a case of a rare disease and assess the treatment planning processes employed by Deepseek-R1, ilustrated in Supplementary Figure~\ref{fig:treament_planing_case1}. We focus on analyzing the differences between the reasoning generated in the formal answers~(for simplicity, denoted as Chain-of-thought) and the default thinking process generated by Deepseek-R1. Our analysis is grounded in evaluating the efficiency of reasoning, the factuality of each step, and the completeness.

First, let’s provide a brief overview of the case. The patient is a 68-year-old male diagnosed with cutis laxa Ehlers-Danlos syndrome (dEDS). His main symptoms include extremely fragile skin, hypermobile joints, and high blood pressure. He has a known allergy to sulfonamides, which previously caused a systemic rash. Regarding his medication status, he was prescribed Triamterene-hydrochlorothiazide (37.5 mg/25 mg daily) to manage his hypertension. Additionally, he has a family history of chronic kidney disease.

\textbf{Treatment Planning Accuracy:}
In terms of the final treatment planning, Deepseek-R1 provided a rational treatment plan. In this case, Deepseek-R1’s output was largely consistent with the ground truth. However, it introduced two novel treatment recommendations: first, to discontinue Triamterene-hydrochlorothiazide and initiate Lisinopril for hypertension management; second, to use Paracetamol for alleviating osteoarthritis pain. Our evaluation pipeline, which integrates a search engine to gather relevant information, confirmed that these recommendations are reasonable. 

Considering the patient’s family history of chronic kidney disease, switching to Lisinopril is a prudent choice, as it aligns with clinical guidelines for managing hypertension in such contexts. Additionally, given the patient’s known allergy to sulfonamides, discontinuing Triamterene-hydrochlorothiazide is justified due to the potential for cross-allergic reactions. Furthermore, the recommendation to use Paracetamol for osteoarthritis pain is appropriate, as it is a commonly accepted treatment option for this condition. 

Therefore, based on these evaluation findings, the final judgment is that the treatment plan proposed by Deepseek-R1 is correct and clinically sound.

\textbf{Efficiency:}
We investigated the effectiveness in the chain-of-thought and the thinking process reasoning. Overall, the output of chain-of-thought is more concise, while that of the thinking process is more fragmented. From the cases, we can see that the thinking process divides the reasoning of the entire problem into 9 steps, whereas Deepseek-R1 only used 5 steps. Additionally, in the first step, the thinking process initially summarized the entire medical record without beginning the reasoning. This is considered an ineffective reasoning step. Therefore, the efficiency of the thinking process is slightly lower, at only 0.89, while the efficiency of chain-of-thought is 1.

\textbf{Factuality:}
In this case, the chain-of-thought and thinking process are largely true and logically sound, with each step building on relevant clinical considerations. The recommendations for managing dEDS, including prioritizing skin protection, dermatology referrals, and gentle wound care protocols, are appropriate given the patient's skin fragility. The approach to osteoarthritis management, which avoids NSAIDs due to the risk of bruising and gastrointestinal bleeding, is also reasonable, especially considering the patient's specific condition. The suggestion to switch hypertension medication away from HCTZ due to the patient's sulfa allergy is cautious and aligns with the potential risk of cross-reactivity, even though the patient has tolerated it thus far. While some plans may be overly cautious or lack specificity, such as the broad suggestion to monitor overall health, the overall reasoning is factually correct and follows a logical progression.

\textbf{Completeness:}
In this case, we find that Deepseek-R1’s thinking process is more complete than the chain-of-thought. The thinking process covers most of the ground-truth rationale, while the chain-of-thought misses some key thinking procedures during its process. Specifically, the chain-of-thought does not explicitly address the overarching objectives of managing dermatosparaxis-type Ehlers-Danlos Syndrome (dEDS), including the comprehensive management of dermatologic, orthopedic, and cardiovascular complications (<Step 1>). Additionally, it fails to outline the treatment plan’s goals of improving physical function, preventing joint injuries and bruising, and maintaining cardiovascular health (<Step 2>). The chain-of-thought also neglects the management of cardiovascular risks associated with vessel fragility and aortic dilation, such as regular imaging surveillance and blood pressure control (<Step 5>), and does not emphasize the importance of multidisciplinary care for specialized support, effective symptom management, and overall quality-of-life improvement (<Step 6>). This leads to a relatively low completeness score.

\subsection{Prompt Collection}
\begin{prompt}
\label{prompt:prompt1_classify_casereport_category}
\textbf{Prompt to classify case reports into "Diagnosis" or "Treatment Planning."}\newline

You are an experienced doctor. I will provide you with the title, abstract, and discussion of a case report. Please help me analyze whether the case report primarily focuses on the diagnostic/treatment process. If the report centers on diagnosis/treatment planning, please respond with "Yes." If the report does not primarily address the diagnostic/treatment planning task, please respond with "No."
\end{prompt}

\begin{prompt}
\label{prompt:prompt2_generate_diagnosis_case}
\textbf{Prompt for generating diagnosis data from case report.}\newline

As an experienced physician, you will receive a patient case report focused on diagnosis. Your task is to:\newline
- Summarize the key information of the patient for diagnosis.\newline
- Summarize the differential diagnosis process, including the rationale for each step and the reasons for considering or excluding specific diagnoses.\newline
- Summarize the final diagnosis of the patient.\newline
Ensure that your summaries are concise and accurate, based solely on the information provided in the case report. Please avoid referencing any images, tables, or other visual data (if any), as that data is no longer accessible.\newline
If the case report is incomplete or does not meet the requirements for summarization, simply output: "I can't."\newline

Format to follow:

\#\#\# Case Summary
Please provide a detailed medical history of the patient, including chief complaint, history of present illness, past medical history, family history, physical examination, results of ancillary tests, and other necessary information for the final diagnosis in the following format.

- Patient Information:\newline
- Chief Complaint: If none, write "None."\newline
- History of Present Illness: If none, write "None."\newline
- Past Medical History: If none, write "None."\newline
- Personal History: If none, write "None."\newline
- Family History: If none, write "None."\newline
- Physical Examination: If none, write "None."\newline
- Ancillary Tests: If none, write "None."\newline

\#\#\# Differential Diagnosis Process\newline
- Firstly, based on the patient's chief complaint and clinical information, an initial list of potential diseases should be generated. This list will then be systematically evaluated by comparing each disease with the patient's symptoms, signs, and test results. \newline
- Then the rationale for including or excluding each diagnosis will be explained based on clinical evidence, highlighting the selection of diagnostic tests that help confirm or rule out specific conditions. The process will involve step-by-step exclusion of less likely diagnoses, ultimately leading to the most probable diagnosis, which will be justified by the clinical reasoning and evidence supporting it.\newline

\#\#\# Final Diagnosis~(with explanation details)\newline
Integrate the patient's clinical presentation, test results, and differential diagnosis process to summarize the final diagnosis. Briefly explain the basis for the diagnosis and highlight the key factors supporting this conclusion.\newline

\#\#\# Diagnosis results \newline
Just Output the diagnostic result without any other explanation.

\end{prompt}

\begin{prompt}
\label{prompt:prompt3_generate_treatmentplanning_case}
\textbf{Prompt for generating treatment planning data from case report.}\newline

As an experienced physician, you will receive comprehensive patient information for treatment planning. Your task is to:\newline
- Summarize provided clinical data, focusing on elements critical for treatment planning. Avoid any information after the treatment.\newline
- Summarize the treatment plan for the patient without revealing any information about the treatment outcomes.\newline
- Avoid referencing any images, tables, or other visual data (if any), as that data is no longer accessible.\newline
Ensure all conclusions are strictly based on the information provided. Do not infer or generate additional information beyond what is given. If the data is insufficient, output: "I can't."\newline

Required Format:\newline
\#\#\# Comprehensive Patient Profile\newline
Organize medical information. Never involve any information after the treatment such as treatment results to avoid knowledge leakage. Also, avoid any information of treatment advice. Just state the patient case. The following is a template you can use to organize the information. You can add or remove content as needed.\newline

- Patient Demographics: [Age/Sex/Other identifiers]\newline
- Chief Complaint: \newline
- History of Present Illness: (including current medication or surgical conditions ...)\newline
- Past Medical History: (including surgical history, medication history ...) \newline
- Allergies: \newline
- Family History: \newline
- Social History:\newline
- Physical Exam: \newline
- Lab Events:\newline
- Imaging Events: \newline
- Final Diagnostic Results: \newline

\#\#\# Treatment Planning Analysis\newline
Summarize the primary objectives of the treatment plan based on the patient's condition and needs. Then, summarize the comprehensive rationale for selecting the preferred treatment plan prior to implementation.\newline
- Object: Objectives of the treatment plan\newline
- Rationale: Combined with the Comprehensive Patient Profile above, explain the reasons for choosing this treatment plan\newline

\#\#\# Selected Treatment for the Patient\newline
Output the treatment in a sentence directly. Use imperative sentences and avoid past tense.\newline

\end{prompt}

\begin{prompt}
\label{prompt:prompt4_body_system_classification}
\textbf{Prompt for body system classification}\newline

Please categorize the body parts involved in the health conditions and disease phenotypes discussed in the following case report. Use the provided title, abstract, and discussion sections to determine the most relevant category. If the body part is not listed, please output "Others." Output the category directly without any additional words. If there are multiple categories, please use \textbackslash n to separate them.\newline
Categories:\newline \newline
Blood, Heart and Circulation\newline
Bones, Joints and Muscles\newline
Brain and Nerves\newline
Digestive System\newline
Ear, Nose and Throat\newline
Endocrine System\newline
Eyes and Vision\newline
Immune System\newline
Kidneys and Urinary System\newline
Lungs and Breathing\newline
Mouth and Teeth\newline
Skin, Hair and Nails\newline
Male Reproductive System\newline
\end{prompt}

\begin{prompt}
\label{prompt:prompt4_disorder_system_classification}
\textbf{Prompt for disorder and condition classification}\newline

Please categorize the diseases and symptoms discussed in the following case report. Use the provided title, abstract, and discussion sections to determine the most relevant category from the list below. If the diseases and symptoms do not fit into any of the listed categories, please output "Others." Output the category directly without any additional words.\newline

Categories:\newline\newline
Cancers\newline
Diabetes Mellitus\newline
Genetics/Birth Defects\newline
Infections\newline
Injuries and Wounds\newline
Mental Health and Behavior\newline
Metabolic Problems\newline
Poisoning, Toxicology, Environmental Health\newline
Pregnancy and Reproduction\newline
Substance Use and Disorders\newline
\end{prompt}

\begin{prompt}
\label{prompt:prompt5_check_if_rare_disease}
\textbf{Prompt used to check whether a certain rare disease are mentioned in the patient case.}\newline

You are an experienced physician. You will be provided with the name of a rare disease, as well as the title, abstract, and Discussion section of a case report. Your task is to determine whether the case report is related to this rare disease. If it is related, output "YES" directly. If it is not related, output "NO" directly.
\end{prompt}

\begin{prompt}
\label{prompt:prompt6_dynamic_diagnosis_gpt4o_giving_info}
\textbf{Prompt designed for GPT-4o to role-play as the patient and provide the required information.}\newline

You are a medical expert providing guidance to a junior physician on a patient case. The junior physician will ask you for additional diagnostic information based on the patient's case details and any available ancillary test results. Your role is to provide accurate and relevant responses regarding the availability of specific diagnostic information.\newline
Guidelines:\newline
1. You will receive the patient's case information and any relevant ancillary test results.\newline
2. The junior physician will ask questions about additional diagnostic information needed for the case.\newline
3. If there is relevant ancillary test information available for the requested diagnostic area, provide the details 4. accurately.\newline
4. If there is no relevant ancillary test information available for the requested diagnostic area, simply state: "There is no relevant ancillary test information available for this request."\newline
Patient Case\newline
\{case\}\newline

Ancillary Test Results\newline
\{ancillary\_test\_results\}

Example Interaction:\newline
Junior Physician: "Does the patient have any imaging studies like an X-ray or CT scan?"\newline
Your Response:\newline
If there is relevant imaging information available:\newline
"Based on the available ancillary test results, the patient has undergone a chest X-ray which shows [specific findings]."\newline
If there is no relevant imaging information available:\newline
"There is no relevant ancillary test information available for this request."\newline
Note: Your responses should be factual and based solely on the provided patient case information and ancillary test results. Avoid speculation or hypotheticals unless explicitly requested.
\end{prompt}

\begin{prompt}
\label{prompt:prompt7_askinfo_dynamic_diagnose}
\textbf{Prompt for asking information under 1-turn examination recommendation setting. } \newline

Please thoroughly examine the patient case summary presented below. Your objective is to perform a detailed diagnostic analysis utilizing all available information. Note that due to the potentially limited details, the preliminary diagnosis may encompass several possible conditions. Should you ascertain that the provided data is inadequate for a definitive conclusion, please enumerate any additional diagnostic tests or information that would be necessary. However, if you can deduce a conclusive diagnosis, please proceed to provide it. Too many requests for information are also inappropriate.\newline

Patient Case Summary:\newline
\{case\}\newline

Guidelines:\newline
Evaluate the patient's symptoms, medical history, and all pertinent details from the case summary.\newline
Formulate differential diagnoses based on your analysis.\newline
If the information is not sufficient for a conclusive diagnosis, specify the further tests or details required.\newline

Always following the response format in each turn of the dialogue, never change the section of \#\#\# format: \newline

\#\#\# Chain of Thought:\newline
[Please sort out your thinking process step by step, with each logical step in a separate paragraph, and use a format such as <step 1> to label each step.]\newline
<step 1> Specific thinking content of this step\newline
<step 2> Specific thinking content of this step\newline
...\newline
<step n> Specific thinking content of this step\newline

\#\#\# Conclusion:\newline
[Give a preliminary conclusion if possible, or summarize the current findings.]\newline

\#\#\# Additional Information Required:\newline
[Indicate if further information is needed by specifying the required tests or data. If a conclusive diagnosis has been made and no additional information is necessary, only output "Not required." directly without any other words in this section.]\newline
For example:\newline
Not required.\newline

or\newline

1. Laboratory tests: details\newline
2. Imaging: details\newline
\end{prompt}

\begin{prompt}
\label{prompt:prompt8_make_final_diagnosis_dynamic_diagnose}
\textbf{Prompt for making the diagnostic decision after 1-turn examination recommendation.} \newline

Please make a final diagnosis for the patient in light of the additional information provided below.\newline

Additional Information:\newline
\{additional\_information\}\newline

Guidelines:\newline
- Evaluate the patient's symptoms, medical history, and all pertinent details from the case summary.\newline
- Formulate differential diagnoses based on your analysis.\newline

Always following the response format in each turn of the dialogue, never change the section of \#\#\# format: \newline

\#\#\# Chain of Thought:\newline
[Please sort out your thinking process step by step, with each logical step in a separate paragraph, and use a format such as <step 1> to label each step.]\newline
<step 1> Specific thinking content of this step\newline
<step 2> Specific thinking content of this step\newline
...\newline
<step n> Specific thinking content of this step\newline

\#\#\# Conclusion:\newline
[Directly output the diagnostic result without any other explanation.]\newline

\end{prompt}

\begin{prompt}
\label{prompt:prompt7_first_turn_free_turn_dynamic_screening}
\textbf{Prompt for the first turn under the free-turn examination recommendation setting.}

Please thoroughly examine the patient case summary presented below. Your objective is to perform a detailed diagnostic analysis utilizing all available information. Note that due to the potentially limited details, the preliminary diagnosis may encompass several possible conditions. Should you ascertain that the provided data is inadequate for a definitive conclusion, please enumerate any additional diagnostic tests or information that would be necessary. However, if you can deduce a conclusive diagnosis, please proceed to provide it. Too many requests for information are also inappropriate.\newline

Patient Case Summary:\newline
\{case\}\newline

Guidelines:\newline
Evaluate the patient's symptoms, medical history, and all pertinent details from the case summary.\newline
Formulate differential diagnoses based on your analysis.\newline
If the information is not sufficient for a conclusive diagnosis, specify the further tests or details required.\newline

Always following the response format in the following dialogue, never change the section of \#\#\# format: \newline

\#\#\# Chain of Thought:\newline
[Please sort out your thinking process step by step, with each logical step in a separate paragraph, and use a format such as <step 1> to label each step.]\newline
<step 1> Specific thinking content of this step\newline
<step 2> Specific thinking content of this step\newline
...\newline
<step n> Specific thinking content of this step\newline

\#\#\# Additional Information Required:\newline
[Indicate if further information is needed by specifying the required tests or data. If a conclusive diagnosis has been made and no additional information is necessary, only output "Not required." directly without any other words in this section.]\newline
For example:\newline
Not required.\newline

or\newline

1. Laboratory tests: details\newline
2. Imaging: details\newline

\#\#\# Conclusion:\newline
[If do not require additional information, please provide a final conclusive diagnosis. Otherwise, summarize the current findings.]\newline
\end{prompt}

\begin{prompt}
\label{prompt:prompt8_subsequent_turns_free_turn_dynamic_screening}
\textbf{Prompt for subsequent turns under free-turn examination recommendation setting.}\newline

Here is the additional information you required. Please proceed with the analysis.\newline

Additional Information:\newline
\{additional\_information\}\newline

Always following the response format in each turn of the dialogue, never change the section of \#\#\# format: \newline
\#\#\# Chain of Thought:\newline
[Please sort out your thinking process step by step, with each logical step in a separate paragraph, and use a format such as <step 1> to label each step.]\newline
<step 1> Specific thinking content of this step\newline
<step 2> Specific thinking content of this step\newline
...\newline
<step n> Specific thinking content of this step\newline

\#\#\# Additional Information Required:\newline
[Indicate if further information is needed by specifying the required tests or data. If a conclusive diagnosis has been made and no additional information is necessary, only output "Not required." directly without any other words in this section.]\newline
For example:\newline
Not required.\newline

or\newline

1. Laboratory tests: details\newline
2. Imaging: details\newline

\#\#\# Conclusion:\newline
[If do not require additional information, please provide a final conclusive diagnosis. Otherwise, summarize the current findings.]\newline

\end{prompt}

\begin{prompt}
\label{prompt:prompt9_accurare_diagnose}
\textbf{Prompt for instructing LLMs to diagnose based on patient case.}\newline

Please carefully study the following patient case summary, conduct a comprehensive and in-depth diagnostic analysis, and clearly provide the final diagnosis result.\newline

\{case\}\newline

Format to Follow:\newline

\#\#\# Reasoning:\newline
[Please sort out your thinking process step by step, with each logical step in a separate paragraph.]\newline
<step 1> Specific thinking content of this step\newline
<step 2> Specific thinking content of this step\newline
...\newline
<step n> Specific thinking content of this step\newline

\#\#\# Answer:\newline
[Just output the diagnostic result without any other explanation.]\newline

\end{prompt}

\begin{prompt}
\label{prompt:prompt10_treament_plan}
\textbf{Prompt for instructing LLMs to devise a treatment plan based on a patient case.}

Please carefully study the following patient case summary, conduct a comprehensive and in-depth treatment planning analysis, and clearly provide the selected treatment for the patient.\newline

\{case\}\newline

Format to Follow:\newline

\#\#\# Chain of Thought:\newline
[Please sort out your thinking process step by step, with each logical step in a separate paragraph, and use a format such as <step 1> to label each step.]\newline
<step 1> Specific thinking content of this step\newline
<step 2> Specific thinking content of this step\newline
...\newline
<step n> Specific thinking content of this step\newline

\#\#\# Answer:\newline
[Just output the selected treatment for the patient without any other explanation.]\newline

\end{prompt}

\begin{prompt}
\label{prompt:prompt11_reformat_unstructured_ground_truth_rationale}
\textbf{Prompt for reformatting the unstructured ground-truth rationale.} \newline
\newline
\# Task Overview

Given a medical case analysis problem with multiple reasoning steps [Text to be organized], reorganize it into clearly structured steps, separated by newline characters.\newline
\newline
\# Organization Requirements\newline
Convert the original solution into a clear, structured reasoning process, while ensuring:

- All original key information is preserved, but if multiple sentences discuss the same topic or serve the same logical reasoning purpose, they can be combined into one reasoning step.

- No new explanations or reasoning are added.

- No steps are omitted.\newline

\# Requirements

- Each step must be atomic (one conclusion per step).

- There should be no content repetition between steps.

- The final answer determination is also considered a step in the logical reasoning.\newline

\# Output Requirements

1.There should be no newline characters within each step, and each step should be separated by a single newline character.

2.For highly repetitive reasoning patterns, output them as a single step.

3.Output a maximum of 10 steps.\newline

\# Output Format

<Step 1> Content of this reasoning step...

<Step 2> Content of this reasoning step...

...

<Step n> Content of this reasoning step...\newline

Below is the text that needs to be reorganized into reasoning steps:

[Text to be organized] \newline

\end{prompt}

\begin{prompt}
\label{prompt:prompt12_effi_classification}
\textbf{Prompt for efficiency classification of each reasoning step.}\newline

\# Task Description

Please analyze and determine the type of the current thinking step based on the provided medical case analysis process, all previous thinking steps, the known patient medical record, and the final reasoning goal. The current thinking step should be classified into one of the following types:

1.Citation: A simple citation, summary, or restatement of information from the patient’s medical record, without generating new reasoning or conclusions.

2.Repetition: A repetition of previous thinking processes, without providing new information or advancing the reasoning process.

3.Reasoning: Providing information beyond what is known, or deriving new conclusions from known information, or proposing new possibilities, which moves the thinking process towards the correct answer and has a direct or indirect effect on the final reasoning goal.

4.Redundancy: Providing new information or possibilities that do not help in reaching the final answer and do not advance the reasoning process.\newline

\# Note

When determining the type, ensure to fully consider the logical relationship and reasoning process between the current thinking step, previous thinking steps, the patient’s medical record, and the reasoning goal. If the current thinking step corresponds to multiple types, select the most appropriate one based on its contribution to the reasoning goal. Maintain objectivity and accuracy in judgment, avoiding subjective assumptions.\newline

\# Output Requirements

Only output your classification of the current thinking step, with possible values being "Citation|Repetition|Reasoning|Redundancy". Do not output any other content.\newline

\# Output Format

[Citation|Repetition|Reasoning|Redundancy]\newline

Now, please classify the following input based on the instructions above:

[Current Thinking Step]\newline

[All Previous Thinking Steps]\newline

[Known Patient Medical Record]\newline

[Final Reasoning Goal]
\end{prompt}

\begin{prompt}
\label{prompt:prompt13_get_keywords}
\textbf{Prompt to extract keywords from the medical text for search engines.}\newline

\# Task Description

You will receive a medical analysis text description [text description to be judged], which involves the analysis and reasoning of a patient’s condition. Your task is to deeply analyze this description, judge the correctness of the medical knowledge described, and the key medical facts supporting the logical reasoning. For this, you need to determine the key medical knowledge points required to make this judgment and provide keywords for retrieving these knowledge points. Ensure your analysis is comprehensive, accurate, and covers all key information in the text description.
Please follow these steps:

1. Carefully read and understand the provided medical analysis text description ([text description to be judged]).

2. Analyze the medical concepts, symptoms, diagnostic methods, treatment principles, etc., involved in the text.

3. Identify the logical reasoning steps in the text and judge whether they are based on correct medical knowledge.

4. Determine the key medical knowledge points required to verify the correctness of these logical reasoning steps.

5. Extract keywords for retrieving these medical knowledge points.\newline

\# Output Requirements

Directly output the list of required keywords, separated by commas, with no other content.\newline

\# Format to Follow

Keyword1, Keyword2,...\newline

Below is the text description to be judged: 

[text description to be judged]
\end{prompt}

\begin{prompt}
\label{prompt:prompt14_check_factuality}
\textbf{Prompt to verify whether each step adheres to  medical priors.}\newline

\# Task Description

You will receive three types of content:

-Patient Case Summary: A summary containing key information about the patient, including basic information, medical history, examination results, etc.

-Text Description to be Judged: A medical text description regarding the patient’s condition, diagnosis, or treatment, which you need to judge for its medical knowledge.

-Known Correct Information: Verified correct medical knowledge related to the above text description, serving as a reference for your judgment.\newline

Your task is to deeply analyze the provided text description[Text Description to be Judged], judge whether its description of the involved medical knowledge is accurate, and whether the key medical facts supporting the logical reasoning are correct. Please proceed as follows:

1.Judge the correctness of the description based on the [Patient Case Summary], [Known Correct Information], and your medical knowledge;

2.If you are still uncertain about a particular description, list search keywords specific to that description. The listed search keywords should be targeted and accurate, helping to further verify the uncertain description.\newline

\# Output Requirements

1.The output format must be valid JSON format, with no other content.

2.Output your judgment in the judgment field, with optional values "Correct$|$Wrong$|$Search".

3.If further search is needed, list the search keywords for the questions you are uncertain about in the keywords\_to\_search field. If the judgment is "Correct" or "Wrong", this field should be "None".\newline

\# Format to Follow:

\texttt{```}\newline

\{\{

"judgment": "Correct|Wrong|Search",

"keywords\_to\_search": "None|keywords",

\}\}

\texttt{```}\newline

Below is the patient information, known correct information, and the text description that needs to be judged for the patient:

[Patient Case Summary]\newline

[Known Correct Information]\newline

[Text Description to be Judged]\newline
\end{prompt}

\begin{prompt}
\label{prompt:prompt15_check_hit}
\textbf{Prompt to verify if the provided reasoning step is included in the generated reasoning process.}\newline

\# Task Overview

Given a reasoning step from a medical case analysis problem, determine whether it appears or is covered in the reference reasoning process.\newline

\# Task Requirements

Evaluate whether the content of the step to be analyzed is the same as or related to any step in the reference reasoning process. Or whether the reference reasoning process covers the core meaning or logic expressed in the reasoning step to be analyzed.\newline

\# Output Requirements

Only output your judgment result on the [Reasoning Step to be Analyzed], with optional values “Yes|No”, do not output any other content.\newline

\# Output Format

[Yes$|$No]\newline

Below is the reasoning step to be analyzed and the reference reasoning process:

[Reasoning Step to be Analyzed]\newline

[Reference Reasoning Process]\newline

\end{prompt}

\begin{prompt}
\label{prompt:prompt9_diagnosis_accuracy_judgement}
\textbf{Prompt for final diagnosis accuracy}\newline

You are a professional medical diagnosis evaluation system. Now, you will receive two diagnosis results: one is the diagnosis predicted by the model ([pred\_diag]), and the other is the verified correct diagnosis ([gt\_diag]). Your task is to judge whether the model-predicted diagnosis([pred\_diag]) is correct.\newline

When evaluating, please consider the following factors:\newline
1.The same disease may have multiple aliases, for example, “Heart disease” may also be called “Cardiac disease”.\newline
2.There may be diversity in language expression, for example, “heart attack” and “myocardial infarction” may refer to the same disease.\newline
3.Only judge whether the diagnosis result is correct, information such as the cause of the disease, symptoms, and treatment recommendations are not included in the evaluation scope.\newline
4.If the correct diagnosis[gt\_diag] is included in the predicted diagnosis but some additional complications are mentioned, it is also considered correct.\newline

\# Output Requirements\newline
Only output your judgment result on the model-predicted [pred\_diag] as “Correct|Wrong”, do not output any other content.\newline

\# Format to Follow:\newline
[Correct|Wrong]\newline

Below is the diagnosis result predicted by the model and the correct diagnosis:\newline
[pred\_diag]\newline
\{pred\_diag\}\newline

[gt\_diag]\newline
\{gt\_diag\}\newline
\end{prompt}

\begin{prompt}
\label{prompt:prompt10_treament_planing_accuracy_judgement}
\textbf{Prompt for treatment plan accuracy}\newline

\# Task Description\newline
As a professional medical treatment planning evaluation system, you will now receive two treatment plan results for assessment: one is the treatment plan predicted by the model ([predicted treatment]), and the other is the verified correct treatment plan ([gt treatment]). Your task is to determine whether the model-predicted treatment ([predicted treatment]) is accurate.\newline

When evaluating, please consider the following factors:\newline
1. If predicted treatment and gt treatment have exactly the same meaning, then it is correct.\newline
2. If the correct treatment plan [gt treatment] is included in the predicted treatment but some additional care are mentioned, it is also considered correct\newline
3. Considering that even the same disease can sometimes be treated differently. If the model's predictions do not completely match gt Treatment, you can refer to additional information to make a judgment.\newline
4. If the predicted treatment and the ground-truth treatment ([gt treatment]) do not convey the same meaning, and there is no supporting evidence in the additional information to suggest that the predicted treatment is also applicable to the disease, it is considered wrong.\newline

\# Output Requirements\newline
Only output your judgment result on the model-predicted [predicted treatment] as “Correct|Wrong”, do not output any other content.\newline

\# Format to Follow:\newline
[Correct|Wrong]\newline

Below is the result predicted by the model and the correct Treatment plan:\newline
[predicted treatment]\newline
\{pred\_treatment\}\newline

[gt treatment]\newline
\{gt\_treatment\}\newline

[Additional Information]\newline
\{additional\_info\}\newline
\end{prompt}

\begin{prompt}
\label{prompt:prompt17_reformat_addtional_infomation_required}
\textbf{Prompt to reformat additional information of auxiliary examinations into structured format.}\newline

\# Task Overview\newline
You will receive an output from an auxiliary diagnostic and treatment large model detailing the additional information required for patient diagnosis analysis [Raw Output Text to be Organized]. Your task is to categorize this content into different information requirement categories, and output it in a JSON dictionary format.\newline

\# Organization Requirements\newline
1. Clearly categorize the original required information output into different information need categories.\newline
2. Each point should clearly indicate:\newline
- "type": The major category of the test item (e.g., laboratory tests, imaging examinations, medical history inquiries, etc.)\newline
- "test\_name":The specific name of the test item (e.g., MRI scan, CT scan, thyroid function test, lump biopsy, etc.)\newline
- "info\_required":The specific content or purpose of the requirement (e.g., to rule out malignant tumors, to better observe the tympanic membrane and middle ear structure, to assess retinal function, etc.)\newline
3. Retain all original content, but if multiple checks are for obtaining the same information or have the same purpose, they can be combined into one point, but the “test\_name” field should include all these checks.\newline
4. Do not add new required content; all information should originate from the original output [raw output text to be organized].\newline
5. Do not omit any steps.\newline

\# Output Requirements\newline
1. The output format must be a valid JSON format without any other content.\newline
2. Output the type of the test item in the “type” field; output the specific name of the test item in the “test\_name” field; output the specific content or purpose of the required information in the “info\_required” field.\newline

\# Output Format\newline
```json\newline
\{\{\newline
"type": "Major Category of the Test Item",\newline
"test\_name": "Specific Name of the Test Item",\newline
"info\_required": "Specific Information Required or the Purpose of the Test"\newline
\}\}\newline
```\newline

Below is the raw output text that needs to be reorganized:\newline

[Raw Output Text to be Organized] \newline
\{info\_required\}\newline

\end{prompt}

\end{document}